\documentclass[twoside]{article}

\usepackage[accepted]{aistats2025}

\usepackage{hyperref}
\usepackage{xcolor}
\hypersetup{
    colorlinks,
    linkcolor={blue!80!black},
    citecolor={blue!80!black},
    urlcolor={blue!80!black}
}

\usepackage{url}            
\usepackage{booktabs}       
\usepackage{amsfonts}       
\usepackage{nicefrac}       
\usepackage{microtype}      
\usepackage{amssymb,amsmath, amsthm, bm, cleveref}
\usepackage{algorithmic}
\usepackage{algorithm}
\usepackage{subcaption}
\usepackage{xcolor}
\usepackage{graphicx}
\usepackage{mathtools}
\usepackage{standalone}
\usepackage{xr}
\begin{document}

\newcommand{\R}{\mathbb{R}}
\newcommand{\x}{\mathbf{x}}
\newcommand{\IL}{\mathbf{I}_{L}}
\newcommand{\z}{\mathbf{z}}
\newcommand{\ahat}{\hat{\mathbf{a}}}
\newcommand{\fp}{\hat\x} 
\newcommand{\y}{\mathbf{y}}
\newcommand{\dx}{\dot{\x}}
\newcommand{\fh}{\hat{f}}
\newcommand{\fnh}{f_{p}}
\newcommand{\fhi}{\fh^i}
\newcommand{\fnhi}{\fnh^i}
\newcommand{\dfdx}{\nabla f(\x)}
\newcommand{\dfdy}{\nabla f(\y)}
\newcommand{\dfdxt}{\nabla f(\xt)}
\newcommand{\dxdt}{\dot{\x}}
\newcommand{\xt}{\x(t)}
\newcommand{\Zeta}{Z}
\newcommand{\Ab}{\mathbf{A}}
\newcommand{\past}{\pi^{\ast}}
\newcommand{\tol}{\delta}


\newcommand{\dataset}{{\cal D}}
\newcommand{\fracpartial}[2]{\frac{\partial #1}{\partial  #2}}

\newcommand{\GJ}[1]{ \textcolor{orange}{\textsc{!GJ:} #1!}}

\newtheorem{assumption}{Assumption}
\newtheorem{theorem}{Theorem}

\newtheorem{lemma}{Lemma}

%

%

\twocolumn[

\aistatstitle{FedECADO: A Dynamical System Model of Federated Learning}

\aistatsauthor{ Aayushya Agarwal \And Gauri Joshi \And  Larry Pileggi }

\aistatsaddress{ Carnegie Mellon University \And  Carnegie Mellon University \And Carnegie Mellon University } ]

\begin{abstract}
Federated learning harnesses the power of distributed optimization to train a unified machine learning model across separate clients. However, heterogeneous data distributions and computational workloads can lead to inconsistent updates and limit model performance. This work tackles these challenges by proposing FedECADO, a new algorithm inspired by a dynamical system representation of the federated learning process. FedECADO addresses non-IID data distribution through an aggregate sensitivity model that reflects the amount of data processed by each client. To tackle heterogeneous computing, we design a multi-rate integration method with adaptive step-size selections that synchronizes active client updates in continuous time. Compared to prominent techniques, including FedProx and FedNova, FedECADO achieves higher classification accuracies in numerous heterogeneous scenarios.
\end{abstract}

\section{Introduction}

Federated learning collaboratively trains machine learning models across distributed compute nodes, each equipped with a distinct local dataset and computational capabilities \cite{mcmahan2017communication,kairouz2021advances}. Employing techniques from distributed optimization, federated learning samples update from individual clients and calculates an aggregate update to refine the global model. This paradigm yields several advantages, such as improving data privacy by keep data at the edge node, improving model generalizability via collaboration with other nodes, and optimizing efficiency when training large datasets.

Nevertheless, compared to distributed optimization, federated learning encounters unique challenges of non-IID data distributions and varying computation capacities amongst local clients. Neglecting the impact of heterogeneity can lead to inconsistent models with limited performance \cite{fednova}. However, previous research has often studied federated learning in the context where local clients have identical learning rates and an IID data distribution \cite{ kairouz2021advances, wang2021field}. As a result, new techniques for aggregating local updates require careful consideration that synchronize heterogeneous client updates for model consistency while being computationally efficient without extensive hyperparameter tuning.  


In this work, we design a new consensus algorithm that address the challenges of non-IID data distribution and heterogeneous computing in federated learning. Using previous work in distributed optimization \cite{ecado} named ECADO, we design new heuristics from a continuous-time model of the federated learning process which represents the trajectory of the global model’s state variables by a continuous-time ordinary differential equation (ODE) whose steady state coincides with the local optimum. The continuous-time model of ECADO provides a new perspective on designing methods that address the challenges facing federated learning. Our main contributions are:
\begin{itemize}
    \item \textbf{Aggregate Sensitivity Model}: A first-order sensitivity model of each client update provides faster convergence in the consensus algorithm and reflects the fraction of data for non-IID data distributions amongst clients.
    \item \textbf{Multi-rate integration with adaptive step sizes}: Local updates from clients with heterogeneous computational capabilities are synchronized in continuous-time using a multi-rate numerical integration. Using properties of numerical accuracy, we propose an aggregation method with adaptive step sizes for faster convergence to steady-state. 
\end{itemize}


Compared to prominent methods such as FedProx and FedNova, our combined approach demonstrates faster convergence and achieves higher classification accuracies for training deep neural network models in numerous heterogeneous computing scenarios.

\section{Related Work}

Federated learning has received significant attention in addressing challenges related to heterogeneous computation and communication overhead. A comprehensive survey of these methods is available in \cite{zhu2021federated,kairouz2021advances,wang2021field}. We specifically focus on methods that target heterogeneous client computation and non-IID data distributions. 

Federated learning methods traditionally rely on discrete iterative algorithms, such as FedAvg \cite{li2020federated} which uses SGD for client training and averages the updated results in each central agent step. While FedAvg offers theoretical guarantees, its performance is often limited in settings with heterogeneous computation and non-IID data distribution \cite{zhao2018federated}. To address these limitations, several modifications have been proposed, including adjustments to the SGD update using state information from the central agent \cite{feddyn, pathak2020fedsplit,  karimireddy2020scaffold, li2020federated} and a Newton-like update in FedDANE \cite{li2019feddane}. One notable method, FedProx \cite{li2020federated}, penalizes client updates deviating far from the central agent step. Other variations of FedProx include the following works \cite{feddyn, li2019convergence}.

Certain federated learning methods have improved convergence rates by introducing new step-size routines \cite{li2019convergence, malinovsky2023server, charles2020outsized} or incorporating momentum into client updates \cite{das2022faster, xu2022coordinating, khanduri2021stem}. Adaptive step size selections, as seen in FedYogi, FedADAM, SCAFFOLD \cite{karimireddy2020scaffold} and FedAdaGrad \cite{reddi2020adaptive}, as well as FedExp \cite{jhunjhunwala2023fedexp}, have also been explored; however, these methods generally do not address heterogeneous computational challenges. A notable method, FedNova \cite{fednova}, specifically addresses heterogeneous computation and data scenarios in federated learning by modifying the gradient update to compensate for variations in client local computation.


Our work adopts a new continuous-time formulation of the federated learning process, where the challenges of heterogeneous computation and non-IID data distributions are seen as updates occurring in parallel in continuous time. This enables us to design new methods based on concepts from dynamical system processes. Our continuous-time formulation of optimization is inspired by work on control systems \cite{behrman1998efficient, attouch1996dynamical, polyak2017lyapunov, wilson2021lyapunov} and circuit simulation principles \cite{agarwal2023equivalent, ecado} to design new optimization algorithms. This approach was previously applied to distributed optimization in ECADO \cite{ecado}, but it was not suitable for federated learning because it assumed full client participation. Our work specifically extends ECADO’s ideas to address the distinct challenges posed by federated learning.
\section{Background on ECADO}
In federated learning, $n$ distributed edge devices collectively train a global model. Each device, $i$, has a locally stored dataset, $\mathcal{D}_i$, and coordinates with the central server to update the parameters of a global machine learning model, represented by a vector $\x$. Due to communication and privacy constraints, the raw local data is not transferred to the central server; instead only model updates or gradients are shared. 

Each edge device trains a localized model, where the local objective function $f_i(\mathbf{x})$ is the empirical risk function with respect to the local dataset $\mathcal{D}_i$, defined as
\begin{equation}
    f_i(\mathbf{x}) = \sum_{\xi \in \mathcal{D}_i} \ell(\mathbf{x};\xi),
    \label{eq:local_obj}
\end{equation}
 where $\xi$ is sample index and $\ell(\mathbf{x};\xi)$ is the sample loss function. The central server seeks to minimize global objective, which is the sum of the local objectives:
%
\begin{align}
    \min_\x f(\x) \text{ where } f(\x) = \sum_{i=1}^n f_i(\x).
    \label{eq:objective_function}
\end{align}
%
To address the challenges of heterogeneous computing and non-IID data distribution, our work designs a new federated learning algorithm based on the centralized distributed optimization framework, ECADO. ECADO analyzes the solution trajectory of the global objective as a continuous-time ODE, known as gradient flow:
\begin{align}
    \dot{\x} &= -\nabla f(\x) \label{eq:gd_flow1}\\
     &= -\sum_{i=1}^n \nabla f_i(\x), \;\; \x(0)=\x_0 \label{eq:gd_flow}
\end{align}
where $\x_0$ are the initial conditions. At steady-state, the gradient flow \eqref{eq:gd_flow} reaches a point where $\dot{\x}=0$, implying $
\nabla f(\x)=0$ as per \eqref{eq:gd_flow1}. Thus, the steady-state aligns with a critical point of the objective function \eqref{eq:objective_function}.

\begin{figure}
    \centering
    \includegraphics[width=0.6\linewidth]{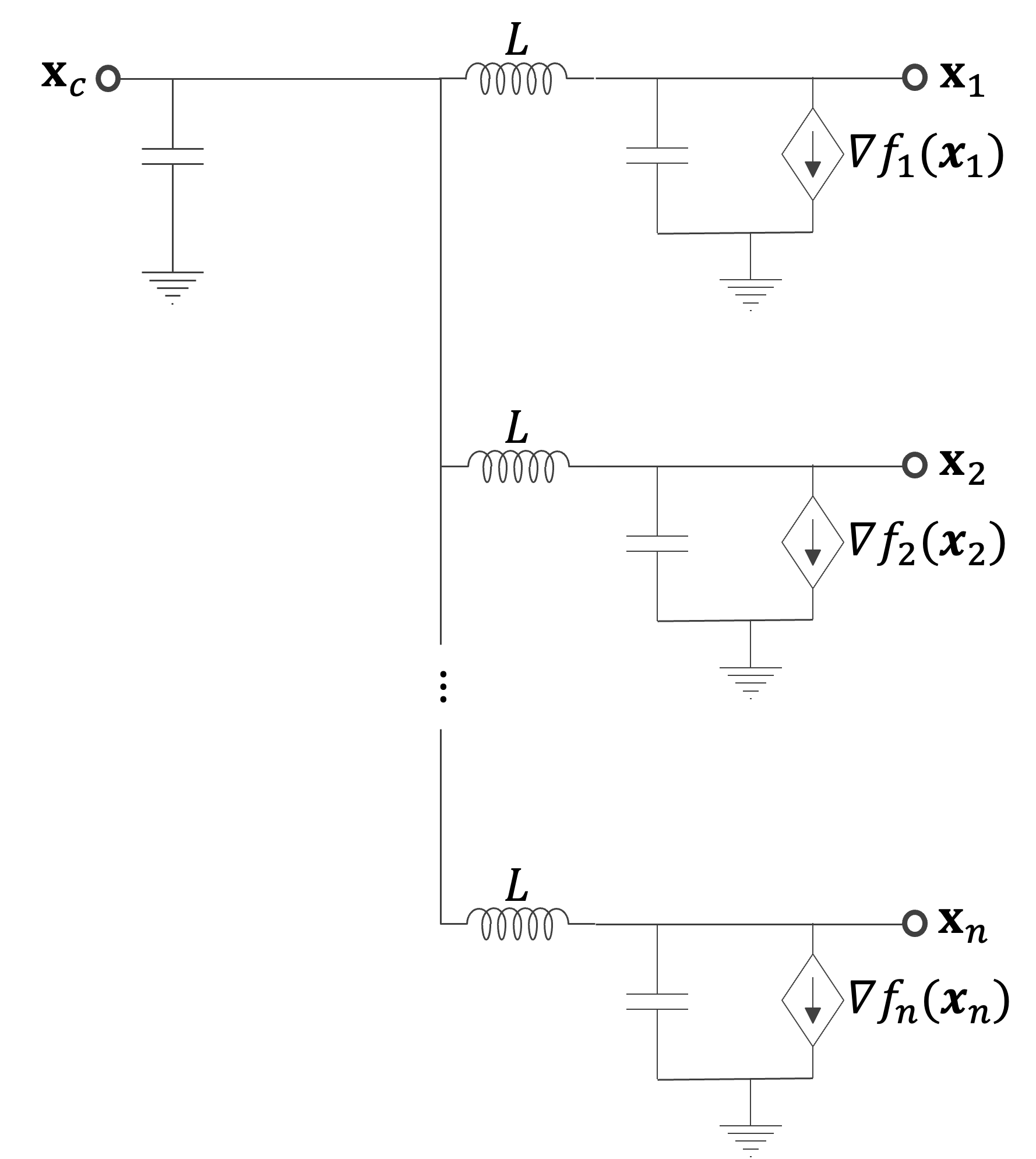}
    \caption{ECADO models federated learning as an equivalent circuit, where node voltages represent state variables, $\x_i$, and gradients, $\nabla f_i(\x_i)$, are voltage-controlled current sources. Using circuit insights, the gradient flow equations \eqref{eq:gd_flow1},\eqref{eq:gd_flow} are modified by introducing an inductor (with an inductance of $L$) between the central agent state, $\x_c$, and the state of each sub-problem, $\x_i$. The resulting gradient flow equations \eqref{eq:central_agent_ode},\eqref{eq:client_ode},\eqref{eq:inductor_ode} are mapped to the equivalent circuit shown. }
    \label{fig:fedecado_circuit}
\end{figure}

Directly solving the gradient flow equations, \eqref{eq:gd_flow1}, to steady-state reveals similar convergence characteristics to those of SGD \cite{ecado}. To improve the convergence of federated learning, ECADO translates the ODE into an equivalent circuit (EC) model, which motivates physics-based optimization techniques that can effectively address the challenges in federated learning. In the EC model depicted in Figure \ref{fig:fedecado_circuit}, the node voltages correspond to the state variables for the central agent and local sub-problems, while the local gradients, $\nabla f_i(\x_i)$, are represented by voltage-controlled current sources. By applying principles from circuit analysis and simulation, we create circuit-inspired algorithms that shape the solution trajectory and select appropriate step sizes.

To separate the central agent state, $\x_c$, from the local state of each subproblem, $\x_i$, in the EC model, ECADO first modifies the gradient flow equations \eqref{eq:gd_flow1},\eqref{eq:gd_flow} by introducing an intermediate flow variable (representing an inductor current in Figure \ref{fig:fedecado_circuit}), $\IL^i$. The flow variables, $\IL^i$, interact with the ODEs of the central agent and local agents according to Kirchhoff's current law (KCL) of the EC as follows:
%
 \begin{align}
     \dot{\x}_c(t) = \sum_{i=1}^n \IL^i(t)  \label{eq:central_agent_ode}\\
    \IL^i(t) + \dot{\x}_i(t) + \nabla f_i(\x_i(t)) =0 \label{eq:client_ode}
\end{align}
Inspired by the current-voltage relationship of an inductor, the flow variables, $\IL^i$, couple the central agent to the local state variables according to:
\begin{align}
        L \dot{\IL}^i (t)= \x_c(t) - \x_i(t) \label{eq:inductor_ode} 
\end{align} 
The flow variable, $\IL^i$, represents the cumulative error between the central agent state, $\x_c$, and local state, $\x_i$, over the simulation window $[0,t]$. This acts as an integral controller for the dynamical system and achieves a second-order effect for faster convergence to steady-state \cite{ecado}.  At steady-state (i.e., critical point in the optimization function), the system reaches an equilibrium where the flow variables, $\IL^i$, are stationary, indicating that $\x_c=\x_i$ for all $i$. The settling time for the continuous-time response of the flow variables, $\IL^i$, is influenced by the hyperparameter $L$ and can be tuned to provide fast convergence as shown in \cite{ecado}.

The modified ODEs \eqref{eq:central_agent_ode}-\eqref{eq:inductor_ode} describes a set of differential equations with the states of all local subproblems, $\x_i$, implicitly coupled. In federated learning, the differential equations are solved over a set of distributed compute nodes. To decouple the ODEs, ECADO proposes an iterative Gauss-Seidel (G-S) method to separate each subproblem from the central agent by treating the intermediate flow variable as constant from the prior iteration. Analyzing the distributed computation as a G-S enables us to study the continuous-time convergence of the full set of ODEs without the effect of discrete updates due to client participation. 

The G-S process separately solves each client independently and subsequently communicates updates the coupling variables at each iteration.
During the ($k+1$)-th iteration of G-S, a client first simulates its local sub-problem (i.e., gradient flow) as follows:
\begin{equation}
     \dot{\x}_i^{k+1}(t) + \nabla f_i(\x_i^{k+1}(t)) + \IL^{i^k}(t) = 0,
\end{equation}
where $\IL^{i^k}$ is the intermediate flow variable that couples the local client ODE to the central agent. In the G-S iteration, $\IL^{i}$ is modeled as a constant from the previous G-S iteration (as indicated by $\IL^{i^k}$). The differential equation is solved using a numerical integration method over a time window of $[t_0,t_1]$. For example, a Forward Euler integration (equivalent to gradient descent \cite{agarwal2023equivalent}) solves for the state at each discrete time point:
\begin{equation}
    \x_i^{k+1}(t+\Delta t) = \x_i^{k+1}(t) - \Delta t ( \nabla f_i(\x_i^{k+1}(t)) + \IL^{i^k}(t)) \label{eq:subproblem_fe}
\end{equation}
 where $\Delta t$ is the time step (or learning rate).

During each iteration of G-S, the local subproblem is simulated for a number of time steps, corresponding to a number of iterations denoted as $e_i$.  Afterwards, each client communicates its local states to the central agent. Using the local client updates, the central agent then updates the flow variables, $\IL^i$, and central agent state, $\x_c$, according to the following ODE:
\begin{align}
    \dot{\x}_c^{k+1}(t) = \sum_{i=1}^n \IL^{i^{k+1}}(t) \label{eq:central_ode1} \\
    L \dot{\IL}^{i^{k+1}}(t) = \x_c^{k+1}(t) - \x_i^{k+1}(t). \label{eq:central_ode2}
\end{align}
In this update, the state variables of each sub-problem is represented by a constant value, $\x_i^{k+1}$, which effectively models the sub-problem for a given time period.

However, the constant-value model of each client assumes that a change in the central agent state, $\x_c$, does not influence $\x_i$ for the time-period. ECADO improves the consensus updates by modeling the first-order effect of each sub-problem due to changes in the state-variable, $\x_c$, using an aggregate sensitivity model of each client's subproblem, denoted by $G_i^{th}$. This improves the convergence of the G-S process captures as we can better capture the coupled interactions between the central agent state and each client. 

$G^{th}$ is defined as the linear sensitivity of the flow variables with respect to a change in the local state, $x_i$:
\begin{align}
    G_i^{th} &= \frac{\partial \IL^i}{\partial \x_i}
\end{align}
The linear sensitivity is calculated using the Hessian of the local objective function as follows:
\begin{align}
    G_i^{th}&= \frac{1}{\Delta t} + \frac{\partial}{\partial \x_i}\nabla f_i(\x_i^{k+1}) 
    \label{eq:sensitivity_derivation}
\end{align}
The full derivation of the sensitivity model is provided in \cite{ecado}. 
However, calculating the Hessian at each G-S iteration is a bottleneck for computation and communication. To reduce the computation load of evaluating the Hessian, ECADO approximates $\frac{\partial}{\partial \x_i}\nabla f_i(\x_i^{k+1})$ with a constant Hessian,
$\bar{H}_i$, which can be periodically computed based on communication and computational capabilities. The constant Hessian is precomputed for each client by averaging the Hessian across a range of samples in the local dataset ($\xi \in\mathcal{D}_i$):
\begin{equation}
    \bar{H} = \frac{1}{|\mathcal{D}_i|}\sum_{\xi\in\mathcal{D}_i} \frac{\partial}{\partial \x_i}\nabla f_{i}(\x_i),
    \label{eq:constant_hessian_def}
\end{equation}
where $\mathcal{D}_i$ is each client's local dataset. 
Using the linear sensitivity, each client update is then represented as:
\begin{equation}
    \IL^{i^{k+1}}(t)G_i^{th^{-1}} + \bar{\x}_i^{k+1}(t),
\end{equation}
where $\bar{\x}_i$ is defined as:
\begin{equation}
   \bar{\x}_i^{k+1}(t) = \x_i^{k+1}(t) - \IL^{i^{k}}(t)G_i^{th^{-1}}.
\end{equation}
The linear sensitivity model is then incorporated into the G-S process as follows:
\begin{equation}
     \dot{\x}_c^{k+1}(t) = \sum_{i=1}^n \IL^{i^{k+1}}(t)     \label{eq:central_ode_thevenin1}
\end{equation}
\begin{equation}
    L \dot{\IL}^{i^{k+1}}(t) = \x_c^{k+1} - (\IL^{i^{k+1}}(t)G_i^{th^{-1}} +\x_i^{k+1}(t) - \IL^{i^{k}}(t)G_i^{th^{-1}}). 
    \label{eq:central_ode_thevenin2}
\end{equation}
The relative values of each client's linear sensitivities drive the central agent states toward certain client updates. ECADO demonstrates that the linear sensitivity model adds a proportional controller to the dynamical system to improve the convergence rate of G-S  \cite{ecado}.

To numerically solve the ODEs describing the dynamics of the central agent states \eqref{eq:central_ode_thevenin1},\eqref{eq:central_ode_thevenin2}, ECADO proposes a Backward Euler (BE) integration step. The BE step is numerically stable and improves the convergence rate of the distributed optimization process \cite{ecado}. The BE step solves for the states at a time $t+\Delta t$:
\begin{equation}
         \x_c^{k+1}(t+\Delta t) = \x_c^{k+1}(t) - \Delta t\sum_{i=1}^{n}\IL^{i^{k+1}}(t+\Delta t) 
         \label{eq:central_ode_be}
\end{equation}
\begin{equation}
\begin{aligned}
     \IL^{i^{k+1}}(t+\Delta t) = \IL^{i^{k+1}}(t) + \frac{\Delta t}{L}  ( \x_c^{k+1}(t+\Delta t) - \\( \IL^{i^{k+1}}(t+\Delta t) G_i^{th^{-1}} + \x_{i}^{k+1}(t+\Delta t) - \IL^{i^k}(t+\Delta t)G_i^{th^{-1}}) ).
     \end{aligned}
     \label{eq:central_ode_be_step}
\end{equation}
This results in the following linear equations:
\begin{equation}
\begin{aligned}
    \begin{bsmallmatrix}
        1+\frac{\Delta t G_1^{th^{-1}}}{L} & 0 & \ldots & -\frac{\Delta t}{L} \\
        0 & 1+\frac{\Delta t G_2^{th^{-1}}}{L} & \ldots & -\frac{\Delta t}{L}\\
        0 & 0 & \ddots & \frac{-\Delta t}{L} \\
        -\Delta t & -\Delta t & \ldots & 1
    \end{bsmallmatrix} \begin{bsmallmatrix}
        \IL^{1^{k+1}}(t+\Delta t) \\ 
        \IL^{2^{k+1}}(t+\Delta t) \\
        \vdots \\
        \x_c^{k+1}(t+\Delta t)
    \end{bsmallmatrix}
    =\\ \frac{\Delta t}{L}\begin{bsmallmatrix}
       -\x_1^{k+1} + \IL^{1^k}(t)G_1^{th^{-1}} \\
       -\x_2^{k+1} + \IL^{2^k}(t)G_2^{th^{-1}} \\
        \vdots \\
        0
    \end{bsmallmatrix} 
    \end{aligned}
    \label{eq:central_agent_be_step2}
\end{equation}
with a global step size of $\Delta t$. At each iteration, ECADO solves \eqref{eq:central_agent_be_step2} to a steady-state defined as $\|\x_c(t+\Delta t) - \x_c(t)\| \leq \tol$, where $\tol$ is the tolerance for convergence.

\section{FedECADO}

While the continuous-time model proposed in ECADO offers fast convergence, it is not directly applicable to federated learning. ECADO's underlying assumption of identically distributed data and synchronized client updates (with identical learning rates and number of epochs) makes it prone to model inconsistencies, as demonstrated in \cite{fednova}. Specifically, ECADO uses a global step-size to perform a BE integration step that solves for consensus step. However, heterogeneous computation prevents synchronized updates between clients because each client model is trained using a different learning rate and number of epochs. 

FedECADO aims to improve the ECADO algorithm to address the challenges of heterogeneous computation and non-IID data distribution in federated learning. 
We propose a mutlirate integration method that synchronizes the client updates to account for heterogenous computation. Additionally, we extend the aggregate sensitivity to model non-IID data distributions.

\subsection{Multi-Rate Integration for Heterogeneous Computation}
\label{sec:multirate_integration_sec}
During each round of communication, a set of active clients transmit their most recent updates to the central agent. The central agent then employs a BE numerical integration technique to compute the states of the central agent in \eqref{eq:multirate_central_ode_thevenin1} and \eqref{eq:multirate_central_ode_thevenin2} based on these local updates.
The BE integration step proposed in ECADO \eqref{eq:central_agent_be_step2} assumes a globally synchronous timescale, where all clients are simulated with the same time step, $\Delta t$, for the same number of epochs. However, this assumption is not applicable to federated learning, where the subset of  actively participating clients, $C_a \in C$,  exhibit a varying step-size, $\Delta t_i$, and number of epochs, $e_i$.

FedECADO tackles this issue by introducing a multirate integration method grounded in a continuous-time perspective of federated learning. We recognize that in continuous time, each active client, $i\in C_a$, simulates its local ODE \eqref{eq:client_ode} for a unique time window, $T_i$:
\begin{equation}
    T_i = \sum_{k=1}^{e_i} \Delta t_i^k.
\end{equation}
This insight builds upon the equivalence between discrete step-sizes and time steps, $\Delta t_i$, resulting in each client essentially simulating its local sub-problem for $e_i$ time steps (i.e., number of epochs). For instance, a client with a local learning rate of $10^{-3}$ and 3 epochs simulates its local ODE for $T_i=3 \times 10^{-3}$ seconds.

\begin{figure*}
\centering
\begin{minipage}{.48\textwidth}
  \centering
    \includegraphics[width=0.6\linewidth]{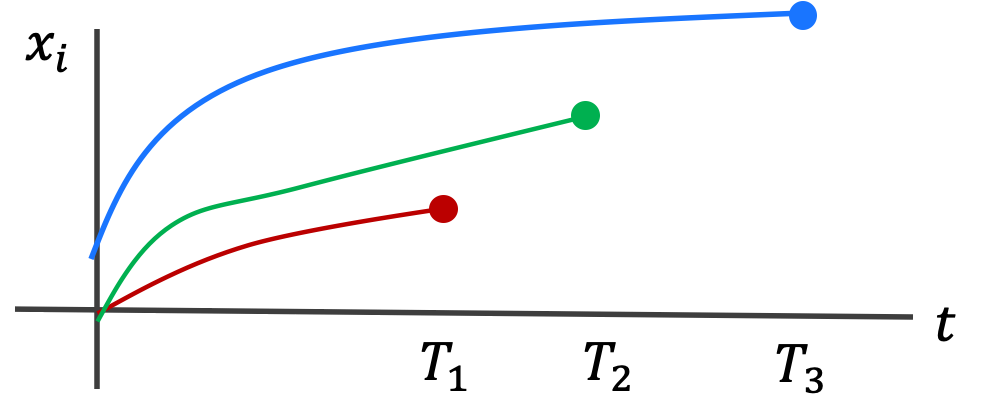}
    \caption{Heterogeneous computation among three clients leads to simulation for different time windows  ($T_1,T_2,T_3$). The final states ($x_1(T_1),x_2(T_2),x_3(T_3)$) are communicated to the central agent, resulting in asynchronous updates.}
    \label{fig:asynchronous_updates}
\end{minipage}%
\hfill
\begin{minipage}{.48\textwidth}
  \centering
\includegraphics[width=0.9\linewidth]{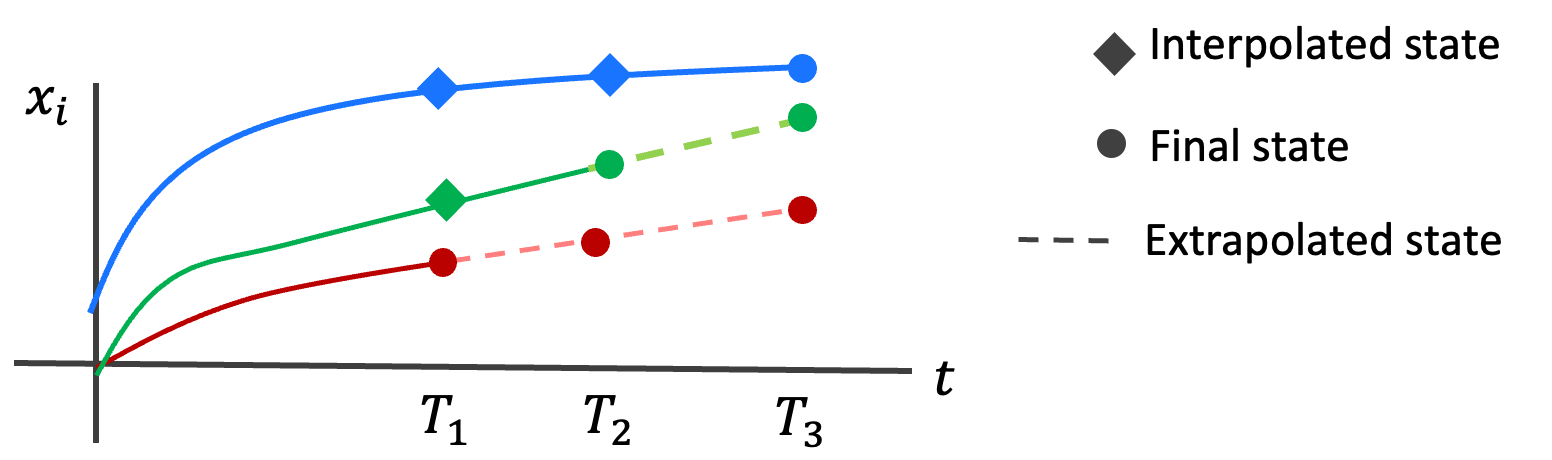}
  \caption{FedECADO proposes a multi-rate integration that evaluates the central agent step at intermediate time points by linearly interpolating and extrapolating client states to the synchronized time point.}
  \label{fig:multirate_integration}
\end{minipage}
\end{figure*}

This continuous-time perspective reveals that each active client simulates its local ODE for a distinct timescale and communicates its final state, $\x_i(T_i)$, to the central agent, leading to an asynchronous update. This issue is illustrated in Figure \ref{fig:asynchronous_updates}, which depicts asynchronous updates from three active clients.
Note, requiring a synchronous timescale is vital for convergence, as all clients reach steady state simultaneously.

\paragraph{Remark 1:}
Convergence to a critical point for the central agent is achieved when all clients simultaneously reach a steady state.

From a continuous-time point of view, Remark 1 illustrates the importance of maintaining a uniform timescale within each sub-circuit to achieve a global steady-state simultaneously.

Inspired by asynchronous distributed circuit simulation \cite{white2012relaxation}, FedECADO introduces a multi-rate integration scheme designed to address the challenge of asynchronous local updates in federated learning. This scheme effectively synchronizes the client updates to ensure accuracy and consistency of the central model.

During each communication round, the multi-rate integration scheme begins by collecting the latest updates from all active clients, $\x_i(T_i) \; \text{ for all } i \in C_a$, along with each client's simulation runtime, denoted by $T_i$. Communicating the simulation time of each client is essential for synchronizing local updates at the central client server and adds minimal computation and communication costs. 
Next, FedECADO solves for the central client states on a synchronous timescale at intermediate timepoints. To synchronize the client updates, we employ a linear interpolation and extrapolation operator, $\Gamma(\x_i(t),\tau)$, that estimates client states, $\x_i(t)$, at an intermediate time point, $\tau$, defined as follows:
\begin{equation}
    \Gamma(\x_i(t),\tau) = \frac{\x_i(t_2)-\x_i(t_1)}{t_2 - t_1} (\tau - t_1)  + \x_i(t_1),
    \label{eq:interpolation_operator}
\end{equation}
where $\x_i(t_2)$ and $\x_i(t_1)$ represent known state values at time points $t_2$ and $t_1$, respectively.

This constructs a synchronous timescale for the central agent to evaluate its state variables for a time window $\tau \in [t_0, t_0+\max(T_i)]$, where $t_0$ is the latest time point in the previous communication round and $\text{max}(T_i)$ is the largest simulation time window for the active clients. The central agent states are now governed by the following ODEs using the operator, $\Gamma(\cdot)$:
\begin{align}
    \dot{\x}_c^{k+1}(\tau) &= \sum_{i=1}^n \IL^{i^{k+1}}(\tau) \label{eq:multirate_central_ode_thevenin1} \\
    L \dot{\IL}^{i^{k+1}}(\tau) &= \x_c^{k+1}(\tau) - (\IL^{i^{k+1}}(\tau)G_i^{th^{-1}} \nonumber\\ &\quad + \Gamma(\x_i^{k+1}(t),\tau) - \IL^{i^{k}}G_i^{th^{-1}}), \label{eq:multirate_central_ode_thevenin2}
\end{align}
where $\Gamma(\x_i^{k+1}(t),\tau)$ calculates the client states estimated at time $\tau$ using the linear interpolation and extrapolation operator \eqref{eq:interpolation_operator}. This operator addresses the challenges posed by asynchronous client updates, which can otherwise lead to model inconsistencies and poor performance. Without the operator, $\Gamma(\x_i^{k+1}(t),\tau)$, the central agent would be forced to incorporate asynchronous client states directly, leading to disparate timescales within the coupled system. This would prevent the central agent and local clients from synchronously reaching a steady state, which has been established by Remark 1 as a necessary condition to converge to a stationary point in the objective function.

FedECADO solves for the central agent states in \eqref{eq:multirate_central_ode_thevenin1},\eqref{eq:multirate_central_ode_thevenin2}  using a numerically stable BE integration method as follows:
\begin{equation}
         \x_c^{k+1}(\tau) = \x_c^{k+1}(\tau) - \Delta t\left(\sum_{i=1}^{n}\IL^{i^{k+1}}(\tau) \right) 
         \label{eq:multirate_central_ode_be}
\end{equation}
\begin{equation}
\begin{aligned}
     \IL^{i^{k+1}}(\tau) = \IL^{i^{k+1}}(t) + \frac{\Delta t}{L}  ( \x_c^{k+1}(\tau) - ( \IL^{i^{k+1}}(\tau) G_i^{th^{-1}} \\+ \Gamma(\x_{i}^{k+1}(t),\tau) - \IL^{i^k}(\tau)G_i^{th^{-1}}) ).
     \end{aligned}
     \label{eq:multirate_central_ode_be_step}
\end{equation}
This results in the following set of linear equations that determine the central agent states at the time-point, $\tau$:
\begin{equation}
    \begin{aligned}
        \begin{bsmallmatrix}
        1+\frac{\Delta t G_1^{th^{-1}}}{L} & 0 & \ldots & -\frac{\Delta t}{L} \\
        0 & 1+\frac{\Delta t G_2^{th^{-1}}}{L} & \ldots & -\frac{\Delta t}{L}\\
        0 & 0 & \ddots & \frac{-\Delta t}{L} \\
        -\Delta t  & -\Delta t  & \ldots & 1
    \end{bsmallmatrix} \begin{bsmallmatrix}
        I_1^{L^{k+1}}(\tau) \\ 
        I_2^{L^{k+1}}(\tau) \\
        \vdots \\
        \x_c^{k+1}(\tau)
    \end{bsmallmatrix}
    =\\ \frac{\Delta t}{L}\begin{bsmallmatrix}
       -\Gamma(\x_{1}^{k+1}(t),\tau) + \IL^{1^k}(t)G_1^{th^{-1}} \\
       -\Gamma(\x_{2}^{k+1}(t),\tau) + \IL^{2^k}(t)G_2^{th^{-1}} \\
        \vdots \\
        0
    \end{bsmallmatrix} 
    \end{aligned}
    \label{eq:central_agent_be_step2}
\end{equation}
Note, $\Delta t$ represents the learning rate for the central agent in federated learning methods and is independent from the client learning rate. 
To establish the convergence properties of the multi-rate integration using the linear interpolation and extrapolation operator, $\Gamma(\cdot)$, we prove that each central agent step in \eqref{eq:multirate_central_ode_be},\eqref{eq:multirate_central_ode_be_step} is a contraction mapping that progressively moves the central agent states toward a stationary point.

\begin{theorem}
The operator $\Gamma(\x,\tau)$, defined in \eqref{eq:interpolation_operator}, synchronizes local client updates and at each evaluation of the central agent states via the FedECADO consensus step in \eqref{eq:central_agent_be_step2} is a contraction mapping towards a stationary point.    
  \label{thm:consensus_convergence}
\end{theorem}

The proof of Theorem \ref{thm:consensus_convergence} is provided in Appendix \ref{sec:consensus_convergence_proof}.



\subsubsection{Selecting Central Agent Step-Size}
During each communication round, we solve the central agent ODEs \eqref{eq:multirate_central_ode_thevenin1},\eqref{eq:multirate_central_ode_thevenin2} using a BE integration. The BE integration is a stable numerical method that approximates the central agent state at time points, $\tau \in [t_0,t_0+\text{max}(T_i)]$ over a time-step, $\Delta t$. We propose adaptively selecting $\Delta t$ using numerical accuracy properties of the BE integration. 

The accuracy of the BE step for the central agent ODEs (\eqref{eq:multirate_central_ode_thevenin1} and \eqref{eq:multirate_central_ode_thevenin2}) can be measured by a local truncation error (LTE) derived in \cite{pillage1998electronic}. The LTE for determining the central agent state, $\varepsilon_{BE}^c$, from \eqref{eq:multirate_central_ode_thevenin1} is estimated as:
\begin{equation}
    \varepsilon_{BE}^C = -\frac{-\Delta t}{2}\left [ \sum_{i=1}^{n}\IL^{i^{k+1}}(\tau) - \sum_{i=1}^{n}\IL^{i^{k+1}}(\tau+\Delta t)  \right ]. \label{eq:be_lte_cap}
\end{equation}

The LTE of the BE integration step for evaluating the flow variables from \eqref{eq:multirate_central_ode_thevenin2}, denoted $\varepsilon_{BE}^L$, is estimated as:
\begin{multline}
    \varepsilon_{BE_i}^L = -\frac{\Delta t}{2L}[ (\x_c^{k+1}(t) - \IL^{i^{k+1}}(t)\bar{G}_i^{th^{-1}} + \x_i^{k+1}(t) - \\ \IL^{i^k}(t)\bar{G}_i^{th^{-1}}) - (\x_c^{k+1}(t+\Delta t) - \IL^{i^{k+1}}(t+\Delta t)\bar{G}_i^{th^{-1}} + \\ \x_i^{k+1}(t+\Delta t) - \IL^{i^k}(t+\Delta t)\bar{G}_i^{th^{-1}})] \label{eq:be_lte_ind}
\end{multline}
where $\bar{G}_i^{th}$ is the sensitivity model derived in \eqref{eq:constant_sensitivity_def}.

To accurately capture the ODE trajectory, we adaptively select the time step to guarantee that the accuracy of the ODE’s BE integration step in \eqref{eq:multirate_central_ode_be},\eqref{eq:multirate_central_ode_be_step} remains within a specified tolerance, $\tol$.
At each iteration, a backtracking line-search style method (shown in Algorithm \ref{adaptive-time-step}) selects a step-size, $\Delta t$, to ensure the following accuracy condition is satisfied:
\begin{equation}
    \text{max} |\varepsilon_{BE} | \leq \tol,
\end{equation}
where $\varepsilon_{BE}=[\varepsilon_{BE}^C, \varepsilon_{BE}^L]$. 

The adaptive time step selection in Algorithm \ref{adaptive-time-step} is initiated by a time step, $\Delta t_0 >0$, which can be selected as a constant hyperparameter or from the previous communication round. A backtracking line-search then adjusts $\Delta t$ to ensure the LTE is bounded by $\tol$. Note, the steady-state convergence in continuous-time guarantees that there exists a $\Delta t>0$ that satisfies the BE accuracy condition, which ensures Algorithm \ref{adaptive-time-step} is bounded \cite{ecado}. Although $\Delta t_0$ is an additional hyperparameter, it does not affect convergence but can influence the number of  backtracking line-search iterations in Algorithm \ref{adaptive-time-step}. 
\begin{algorithm}
    \caption{Adaptive Time Stepping Method}
    \label{adaptive-time-step}
    \textbf{Input: } $L>0, \tol>0, \Delta t_0 >0$
    
    \begin{algorithmic}[1]
    \STATE{$\Delta t \leftarrow \Delta t_0$}
    \STATE{\textbf{do while}  $\max(|\varepsilon_{BE}|) \leq \tol$}
    \STATE{\hspace*{\algorithmicindent}$\Delta t = \frac{\tol}{max(|\varepsilon_{BE})} \Delta t$}
    
    \STATE{\hspace*{\algorithmicindent}Compute \small{$\x_c^{k+1}(\tau +\Delta t), \IL^{i^{k+1}}(\tau +\Delta t)$} using \eqref{eq:central_agent_be_step2}}
     \STATE{\hspace*{\algorithmicindent}Evaluate $\varepsilon_{BE}=[\varepsilon_{BE}^C,\varepsilon_{BE}^L]$ using \eqref{eq:be_lte_cap},\eqref{eq:be_lte_ind}} 
    \RETURN $\Delta t$
    \end{algorithmic}
    \end{algorithm}

\subsection{Aggregate Sensitivity Model}
In federated learning, non-IID data distributions can result in inconsistent models if not addressed during the consensus step. To represent the non-IID data distributions, we extend the aggregate sensitivity in ECADO to account for these variations in the following joint optimization with non-IID data distributions:
\begin{align}
    \min_\x f(\x)  \\
    f(\x) = \sum_{i}^{n} p_i f_i(\x) \label{eq:non_iid_objective}
\end{align}
where $p_i$ scales the contribution of $f_i(\x)$ to the overall objective based on the size of local dataset, $\mathcal{D}_i$, relative to the total dataset size, $\mathcal{D}$, defined as:
\begin{equation}
    p_i = |\mathcal{D}_i|/|\mathcal{D}|.
\end{equation}

The gradient flow equations for the federated learning objective with non-IID data distributions \eqref{eq:non_iid_objective} is:
\begin{equation}
    \dot{\x}(t) = -p_i \nabla f(\x(t)), \;\; \x(0)=\x_0.
\end{equation}
Using the modified gradient flow based on the EC in Figure \ref{fig:fedecado_circuit}, the set of ODEs describing the circuit are:
\begin{align}
    \dot{\x}_c =  \sum_{i\in C}\IL^i(t)\\
    \IL^i(t) + 
    \dot{\x}_i(t)+ p_i \nabla  f_i (\x_i(t)) = 0 \label{eq:client_ode_w_norm} \\
    L\dot{\IL}^i(t) = \x_c(t) - \x_i(t),
\end{align}
where the local gradients are scaled by the relative dataset size, $p_i$, in \eqref{eq:client_ode_w_norm}.

To reflect the varying data distributions of each client in the consensus step, we derive an aggregate sensitivity model for each client that incorporates the relative dataset size, $p_i$, and incorporate it in the G-S process \eqref{eq:central_ode_thevenin1}-\eqref{eq:central_ode_thevenin2} for faster convergence. 
The aggregate sensitivity model for each client is defined as follows:
\begin{equation}
    G_{th}^i = \frac{\partial \IL^i}{\partial \x_i}.
\end{equation}
Using \eqref{eq:client_ode_w_norm}, we evaluate the sensitivity model, $G_{th}^i$, as 
\begin{align}
     G_{th}^i = p_i \frac{\partial}{\partial \x_i} \nabla f_i(\x_i) + \frac{\partial}{\partial \x_i}\dot{\x}_i. \label{eq:sensitivity_model_w_de}
\end{align}

The sensitivity model in \eqref{eq:sensitivity_model_w_de} includes a partial of a time-derivative term. Assuming a BE step for the ODE, we can numerically evaluate $G_{th}^i$ as:
\begin{equation}
     G_{th}^{i} = \frac{1}{\Delta t} + p_i \nabla^2 f_i(\x_i).
\end{equation}
The derivation of $G_{th}^{i}$ is provided in \cite{ecado}. In order to mitigate the computational demands associated with evaluating the Hessian, $\nabla^2 f$, for individual datapoints, FedECADO introduces a constant aggregate sensitivity, $\bar{G}_i^{th}$ , which is derived by averaging the Hessian across a subset of datapoints. Employing a constant value to approximate the Hessian, $\bar{H} \approx \nabla^2 f_i$, defines a constant sensitivity model for each client:
\begin{align}
\bar{G}_{th}^{i} &= \frac{1}{\Delta t} + p_i \bar{H}^i 
\label{eq:constant_sensitivity_def}
\end{align}
 $\bar{G}_{th}^{i}$ scales the constant sensitivity model in \eqref{eq:sensitivity_derivation} by $p_i$ to overcome the limitations in ECADO. This sensitivity model can be periodically updated to reassess each client's first-order response, balancing the trade-off between communication and computation. Intuitively, this approach suggests that a client with a larger local dataset will produce a higher $\bar{G}_{th}^i$, thereby having a greater influence on the central agent's state updates. From the perspective of dynamical systems, a higher $\hat{G}_{th}^i$ introduces a larger proportional controller for the respective client, leading to a relatively quicker convergence towards the corresponding update.

\section{Experiments}

We evaluate FedECADO's performance by training multiple deep neural network models distributed across multiple clients. The full FedECADO workflow is shown in Algorithm \ref{fedecado_alg} and detailed in Section \ref{sec:fedecado_alg}. We benchmark our approach against established federated learning methods designed for heterogeneous computation, including FedProx \cite{li2020federated} and FedNova \cite{fednova}. Our experiments focus on two key challenges: non-IID data distribution and asynchronous client training. Wethen  demonstrate the scalability of FedECADO on larger models with both non-IID data distributions and asynchronous training in Section \ref{sec:scaling_fedecado}. In these scenarios, FedECADO achieves higher classification accuracy and lower training loss, thus demonstrating its efficacy for real-world federated learning applications.

\subsection{Non-IID Data Distribution}
We evaluate FedECADO's performance by training a VGG-11 model \cite{vgg} on the non-IID CIFAR-10 \cite{cifar} dataset distributed across 100 clients. To model real-world scenarios with limited client participation, we set an active participation ratio of 0.1, meaning only 10 clients actively participate in each communication round. The data distribution adheres to a non-IID Dirichlet distribution ($\text{Dir}_{16}(0.1)$). 
 The specific dataset size, $|\mathcal{D}_i|$, for each client is predetermined according to the Dirichlet distribution before training and used to precalculate the average sensitivity model proposed in \eqref{eq:constant_sensitivity_def}. In these experiments, the average sensitivity model is not updated during training.

\begin{figure}
  \centering
  \begin{subfigure}[t]{0.45\columnwidth}
        \centering
        \includegraphics[width=1.2\columnwidth,height=3cm]{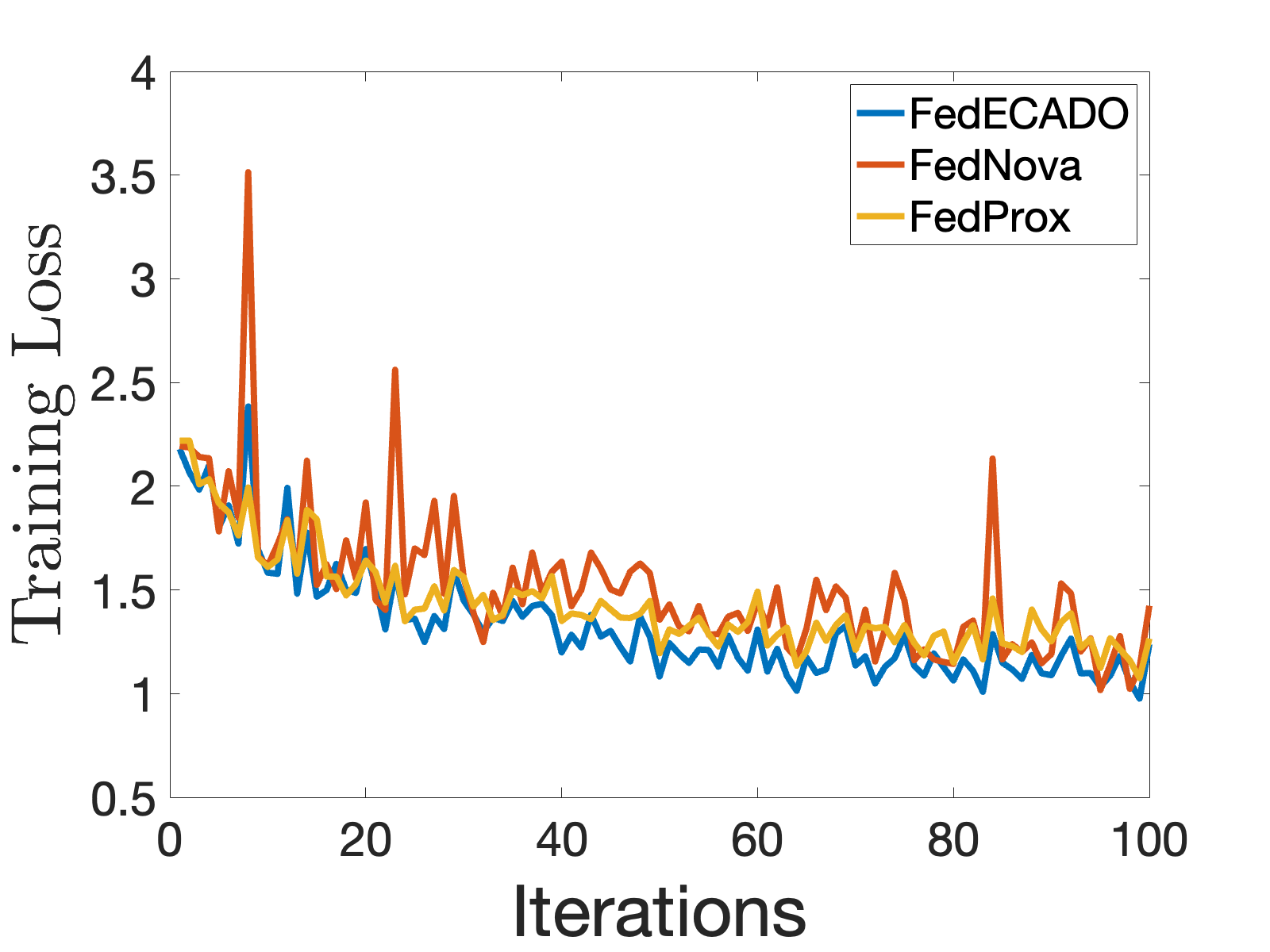}
    \end{subfigure}%
    \hfill
    \begin{subfigure}[t]{0.45\columnwidth}
        \centering
\includegraphics[width=1.2\columnwidth,height=3cm]{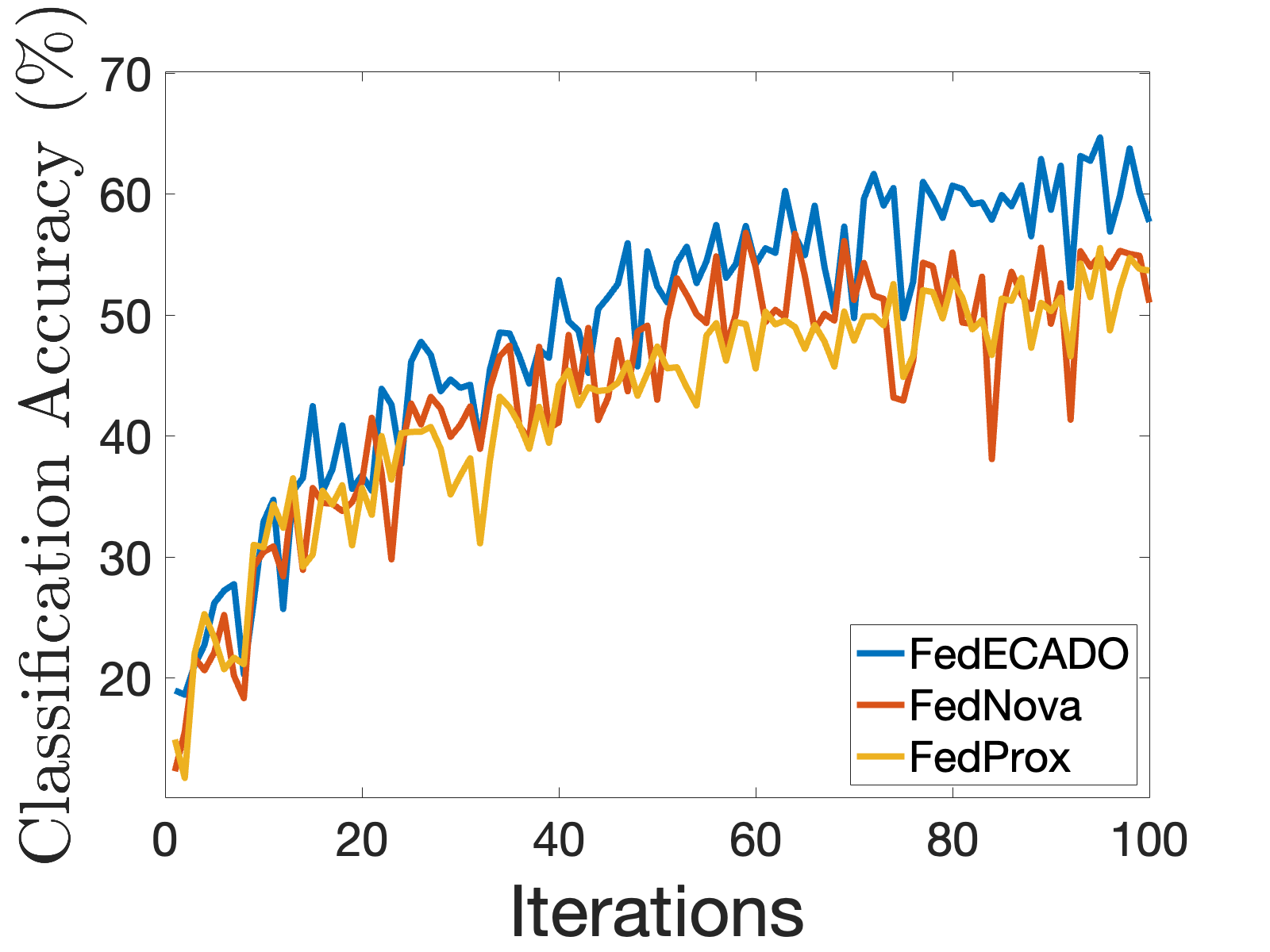}
    \end{subfigure}
    \caption{The training loss and classification accuracy for a VGG-11 model trained on a CIFAR-10 dataset across 100 clients with non-IID Dirichlet distribution.}
    \label{fig:non_iid_test}
\end{figure}

Using each federated learning method, we train for 200 epochs, examining the training loss and classification accuracy at each step. As illustrated in Figure \ref{fig:non_iid_test}, FedECADO achieves the highest classification accuracy throughout the training process (with an improvement of 7\% compared to FedNova and 13\% compared to FedProx). This demonstrates the efficacy of its aggregate sensitivity model in adapting to data heterogeneity.

To test FedECADO's robustness, we repeat the experiment 20 times with different data partitioning defined by Dirichlet distribution. Table \ref{tab:non_iid_acc} shows the mean and standard deviation (std) of each methods' classification accuracies after 200 epochs. FedECADO exhibits the highest mean accuracy with low variance, demonstrating its effectiveness across diverse data distributions.

\begin{figure}
\centering
 \begin{subfigure}[t]{0.45\columnwidth}
        \centering
        \includegraphics[width=1.2\columnwidth,height=3.2cm]{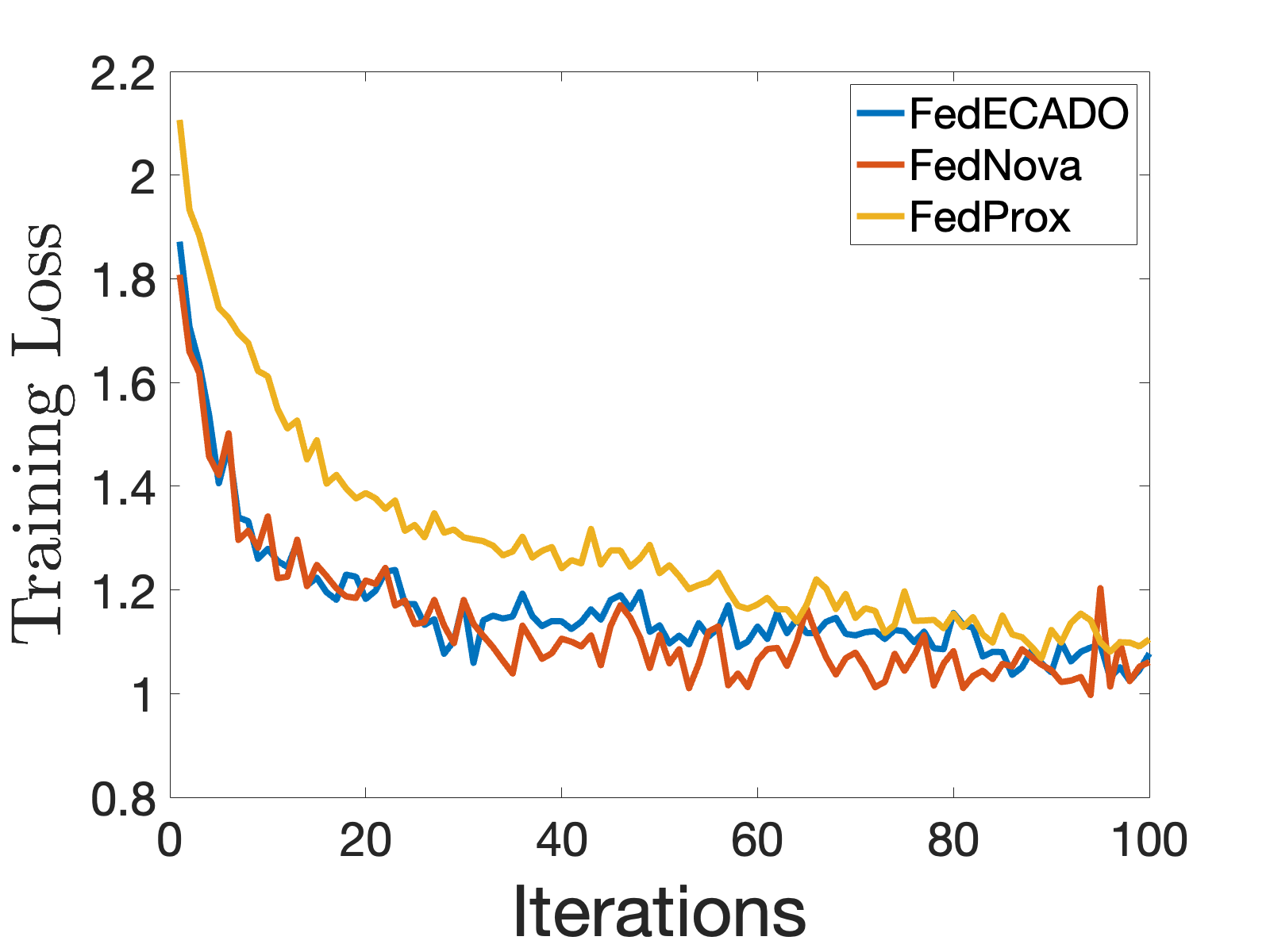}
    \end{subfigure}%
    \hfill
    \begin{subfigure}[t]{0.45\columnwidth}
        \centering
        \includegraphics[width=1.2\columnwidth,height=3.2cm]{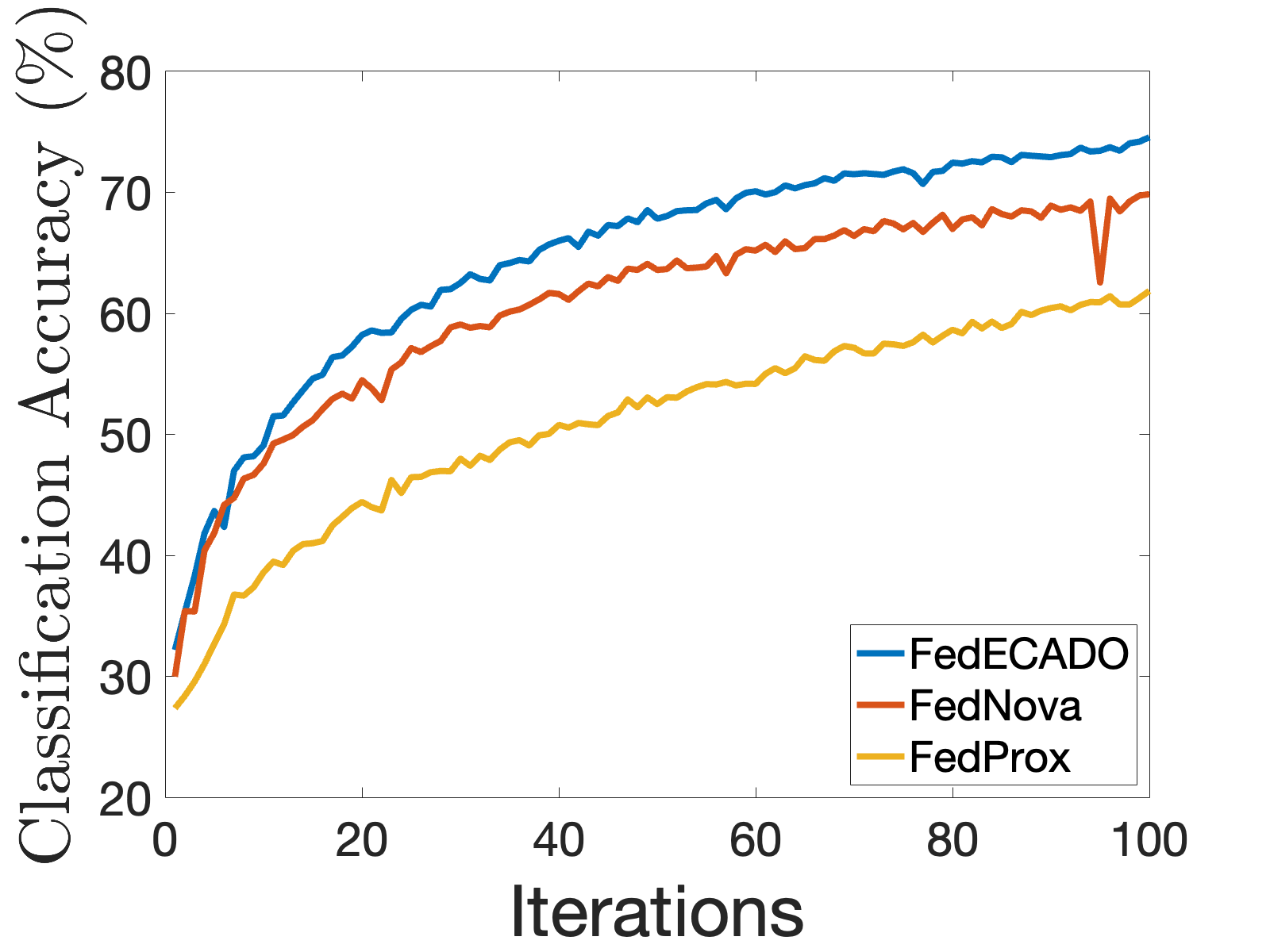}
    \end{subfigure}
    \caption{The training loss and classification accuracy of a VGG-11 model trained on a CIFAR-10 dataset across 100 clients where each client's learning rate and number of epochs is randomly determined by \eqref{eq:random_lr},\eqref{eq:random_epochs}.}
    \label{fig:asynchronous_test}.
\end{figure}

\begin{table}[]
    \centering
    \begin{tabular}{|p{2cm}|c|c|c|}
    \hline
         Classification Acc. (\%)& FedECADO & FedNova & FedProx \\ \hline
        Mean (Std.) & 57.8 (3.6) & 48.9 (2.9) &44.3 (3.2)\\ \hline
    \end{tabular}
    \caption{Classification accuracies for training a VGG-11 model on CIFAR-10 dataset distributed across 100 with Dirichlet data distribution for 200 epochs}
    \label{tab:non_iid_acc}
\end{table}

\subsection{Asynchronous Computation}

In the following experiment, we evaluate the performance of the multi-step integration proposed in Section \ref{sec:multirate_integration_sec}. We train the VGG-11 model \cite{vgg} on a CIFAR-10 dataset \cite{cifar} for 100 epochs across 100 clients with an IID data distribution. However, each client exhibits a different learning rate, $lr_i$, and number of epochs, $e_i$, whose values are sampled by a uniform distribution:
\begin{align}
    lr_i \sim U[10^{-4}, 10^{-3}]     \label{eq:random_lr}
\\
    e_i \sim U[1,10].
    \label{eq:random_epochs}
\end{align}
Figure \ref{fig:asynchronous_test} highlights the training loss and classification accuracy for a single random sample of $lr_i$ and $e_i$ using FedECADO, FedNova, and FedProx.

FedECADO’s multi-rate integration synchronizes the heterogeneous updates of active clients at each communication round, resulting in faster convergence toward a local minimum of the objective and higher classification accuracy after 100 epochs. Note, due to the IID data distribution, the improvement is solely attributed to the multi-rate integration because the aggregate sensitivity model, $\bar{G}^{th}$, is identical for each client.

FedECADO's improvement is further demonstrated across multiple runs, where the learning rate and number of epochs are randomly selected according to \eqref{eq:random_lr},\eqref{eq:random_epochs}. As shown in Table \ref{tab:asynch_acc}, FedECADO achieves a higher mean classification accuracy (compared to FedProx and FedNOVA) and the low variance indicates that it performs well across a range of client settings. We further demonstrate the scalability of FedEcado in heterogeneous settings in Section \ref{sec:scaling_fedecado}.

\begin{table}[]
    \centering
    \begin{tabular}{|p{2cm}|c|c|c|}
    \hline
        Classification Acc. (\%) & FedECADO & FedNova & FedProx \\ \hline
        Mean (Std.) & 72.2 (3.8) & 67.6 (5.2) &61.4 (4.8)\\ \hline
    \end{tabular}
    \caption{Classification accuracies for a VGG-11 model trained on a CIFAR-10 dataset across 100 clients with each client learning rate and epochs set by \eqref{eq:random_lr},\eqref{eq:random_epochs}.}
    \label{tab:asynch_acc}
\end{table}

\section{Conclusion}

We introduce a new federated learning algorithm, FedECADO, inspired by a dynamical system of the underlying optimization problem, which addresses the challenges of heterogeneous computation and non-IID data distribution. To handle non-IID data distribution, FedECADO constructs an aggregate sensitivity model that is integrated into the central agent update for more accurate model adjustments. To address heterogeneous computation in federated learning, FedECADO employs a linear interpolation and extrapolation algorithm that synchronizes client updates at each communication round. The central model state is then evaluated using a new multi-rate integration, which adaptively selects step-sizes based on numerical accuracy, thus guaranteeing convergence to a critical point. We demonstrate the efficacy of FedECADO through distributed training of multiple DNN models across diverse heterogeneous settings. Compared to prominent federated learning methods, FedECADO consistently achieves higher classification accuracies, underscoring its effectiveness in training distributed DNN models with varying client capabilities and data distributions.

\newpage
\bibliographystyle{plain} 
\bibliography{sample_bib}

\begin{thebibliography}{10}

\bibitem{feddyn}
Durmus Alp~Emre Acar, Yue Zhao, Ramon~Matas Navarro, Matthew Mattina, Paul~N Whatmough, and Venkatesh Saligrama.
\newblock Federated learning based on dynamic regularization.
\newblock {\em arXiv preprint arXiv:2111.04263}, 2021.

\bibitem{agarwal2023equivalent}
Aayushya Agarwal, Carmel Fiscko, Soummya Kar, Larry Pileggi, and Bruno Sinopoli.
\newblock An equivalent circuit workflow for unconstrained optimization, 2023.

\bibitem{ecado}
Aayushya Agarwal and Larry Pileggi.
\newblock An equivalent circuit approach to distributed optimization.
\newblock {\em arXiv preprint arXiv:2305.14607}, 2023.

\bibitem{attouch1996dynamical}
Hedy Attouch and Roberto Cominetti.
\newblock A dynamical approach to convex minimization coupling approximation with the steepest descent method.
\newblock {\em Journal of Differential Equations}, 128(2):519--540, 1996.

\bibitem{behrman1998efficient}
William Behrman.
\newblock {\em An efficient gradient flow method for unconstrained optimization}.
\newblock stanford university, 1998.

\bibitem{charles2020outsized}
Zachary Charles and Jakub Kone{\v{c}}n{\`y}.
\newblock On the outsized importance of learning rates in local update methods.
\newblock {\em arXiv preprint arXiv:2007.00878}, 2020.

\bibitem{das2022faster}
Rudrajit Das, Anish Acharya, Abolfazl Hashemi, Sujay Sanghavi, Inderjit~S Dhillon, and Ufuk Topcu.
\newblock Faster non-convex federated learning via global and local momentum.
\newblock In {\em Uncertainty in Artificial Intelligence}, pages 496--506. PMLR, 2022.

\bibitem{jhunjhunwala2023fedexp}
Divyansh Jhunjhunwala, Shiqiang Wang, and Gauri Joshi.
\newblock Fedexp: Speeding up federated averaging via extrapolation.
\newblock {\em arXiv preprint arXiv:2301.09604}, 2023.

\bibitem{kairouz2021advances}
Peter Kairouz, H~Brendan McMahan, Brendan Avent, Aur{\'e}lien Bellet, Mehdi Bennis, Arjun~Nitin Bhagoji, Kallista Bonawitz, Zachary Charles, Graham Cormode, Rachel Cummings, et~al.
\newblock Advances and open problems in federated learning.
\newblock {\em Foundations and trends{\textregistered} in machine learning}, 14(1--2):1--210, 2021.

\bibitem{karimireddy2020scaffold}
Sai~Praneeth Karimireddy, Satyen Kale, Mehryar Mohri, Sashank Reddi, Sebastian Stich, and Ananda~Theertha Suresh.
\newblock Scaffold: Stochastic controlled averaging for federated learning.
\newblock In {\em International conference on machine learning}, pages 5132--5143. PMLR, 2020.

\bibitem{khanduri2021stem}
Prashant Khanduri, Pranay Sharma, Haibo Yang, Mingyi Hong, Jia Liu, Ketan Rajawat, and Pramod Varshney.
\newblock Stem: A stochastic two-sided momentum algorithm achieving near-optimal sample and communication complexities for federated learning.
\newblock {\em Advances in Neural Information Processing Systems}, 34:6050--6061, 2021.

\bibitem{cifar}
Alex Krizhevsky, Geoffrey Hinton, et~al.
\newblock Learning multiple layers of features from tiny images.
\newblock 2009.

\bibitem{li2020federated}
Tian Li, Anit~Kumar Sahu, Manzil Zaheer, Maziar Sanjabi, Ameet Talwalkar, and Virginia Smith.
\newblock Federated optimization in heterogeneous networks.
\newblock {\em Proceedings of Machine learning and systems}, 2:429--450, 2020.

\bibitem{li2019feddane}
Tian Li, Anit~Kumar Sahu, Manzil Zaheer, Maziar Sanjabi, Ameet Talwalkar, and Virginia Smithy.
\newblock Feddane: A federated newton-type method.
\newblock In {\em 2019 53rd Asilomar Conference on Signals, Systems, and Computers}, pages 1227--1231. IEEE, 2019.

\bibitem{li2019convergence}
Xiang Li, Kaixuan Huang, Wenhao Yang, Shusen Wang, and Zhihua Zhang.
\newblock On the convergence of fedavg on non-iid data.
\newblock {\em arXiv preprint arXiv:1907.02189}, 2019.

\bibitem{malinovsky2023server}
Grigory Malinovsky, Konstantin Mishchenko, and Peter Richt{\'a}rik.
\newblock Server-side stepsizes and sampling without replacement provably help in federated optimization.
\newblock In {\em Proceedings of the 4th International Workshop on Distributed Machine Learning}, pages 85--104, 2023.

\bibitem{mcmahan2017communication}
Brendan McMahan, Eider Moore, Daniel Ramage, Seth Hampson, and Blaise~Aguera y~Arcas.
\newblock Communication-efficient learning of deep networks from decentralized data.
\newblock In {\em Artificial intelligence and statistics}, pages 1273--1282. PMLR, 2017.

\bibitem{pathak2020fedsplit}
Reese Pathak and Martin~J Wainwright.
\newblock Fedsplit: An algorithmic framework for fast federated optimization.
\newblock {\em Advances in neural information processing systems}, 33:7057--7066, 2020.

\bibitem{pillage1998electronic}
Lawrence Pillage.
\newblock {\em Electronic Circuit \& System Simulation Methods (SRE)}.
\newblock McGraw-Hill, Inc., 1998.

\bibitem{polyak2017lyapunov}
Boris Polyak and Pavel Shcherbakov.
\newblock Lyapunov functions: An optimization theory perspective.
\newblock {\em IFAC-PapersOnLine}, 50(1):7456--7461, 2017.

\bibitem{reddi2020adaptive}
Sashank Reddi, Zachary Charles, Manzil Zaheer, Zachary Garrett, Keith Rush, Jakub Kone{\v{c}}n{\`y}, Sanjiv Kumar, and H~Brendan McMahan.
\newblock Adaptive federated optimization.
\newblock {\em arXiv preprint arXiv:2003.00295}, 2020.

\bibitem{vgg}
Karen Simonyan and Andrew Zisserman.
\newblock Very deep convolutional networks for large-scale image recognition.
\newblock {\em arXiv preprint arXiv:1409.1556}, 2014.

\bibitem{wang2021field}
Jianyu Wang, Zachary Charles, Zheng Xu, Gauri Joshi, H~Brendan McMahan, Maruan Al-Shedivat, Galen Andrew, Salman Avestimehr, Katharine Daly, Deepesh Data, et~al.
\newblock A field guide to federated optimization.
\newblock {\em arXiv preprint arXiv:2107.06917}, 2021.

\bibitem{fednova}
Jianyu Wang, Qinghua Liu, Hao Liang, Gauri Joshi, and H~Vincent Poor.
\newblock Tackling the objective inconsistency problem in heterogeneous federated optimization.
\newblock {\em Advances in neural information processing systems}, 33:7611--7623, 2020.

\bibitem{white2012relaxation}
Jacob~K White and Alberto~L Sangiovanni-Vincentelli.
\newblock Relaxation techniques for the simulation of vlsi circuits.
\newblock 2012.

\bibitem{wilson2021lyapunov}
Ashia~C Wilson, Ben Recht, and Michael~I Jordan.
\newblock A lyapunov analysis of accelerated methods in optimization.
\newblock {\em Journal of Machine Learning Research}, 22(113):1--34, 2021.

\bibitem{xu2022coordinating}
An~Xu and Heng Huang.
\newblock Coordinating momenta for cross-silo federated learning.
\newblock In {\em Proceedings of the AAAI Conference on Artificial Intelligence}, volume~36, pages 8735--8743, 2022.

\bibitem{zhao2018federated}
Yue Zhao, Meng Li, Liangzhen Lai, Naveen Suda, Damon Civin, and Vikas Chandra.
\newblock Federated learning with non-iid data.
\newblock {\em arXiv preprint arXiv:1806.00582}, 2018.

\bibitem{zhu2021federated}
Hangyu Zhu, Jinjin Xu, Shiqing Liu, and Yaochu Jin.
\newblock Federated learning on non-iid data: A survey.
\newblock {\em Neurocomputing}, 465:371--390, 2021.

\end{thebibliography}

\newpage
\onecolumn

\aistatstitle{Supplementary Materials}
\section{Proof of Theorem \ref{thm:consensus_convergence}}
\label{sec:consensus_convergence_proof}
\begin{proof}
The convergence proof of FedECADO relies on the following assumptions for each local objective function, $f_(\x)$.

\begin{assumption} \label{a1} (Boundedness)
$f\in C^2$ and $\inf_{\x\in\R^n}f(\x)>-R$ for some $R>0$. \label{a1}
\end{assumption}

\begin{assumption}\label{a2}
    (Coercive) $f$ is coercive (i.e., $\lim_{\|\x\|\to\infty} f(\x) = +\infty$)
\end{assumption}

\begin{assumption} \label{a3}
(Lipschitz and bounded gradients): for all $x,y\in\R^n$,  $\Vert \dfdx-\dfdy\Vert\leq L\Vert\x-\y\Vert$, and $\Vert\dfdx\Vert\leq B$ for some $B>0$.\end{assumption}

In order to analyze the convergence of the FedECADO consensus step \eqref{eq:central_agent_be_step2} using the interpolation/extrapolation operator \eqref{eq:interpolation_operator}, we represent all the state variables of the central agent, including the central agent state $\x_c$ and the flow variables $\IL^i$, as a vector $X=[\IL^1, \IL^2,\ddots, \IL^n, \x_c,]$. The ODE for the central agent state in FedECADO, as defined by equations \eqref{eq:multirate_central_ode_be} and \eqref{eq:multirate_central_ode_be_step}, can be generalized as follows: 
\begin{equation} 
\dot{X}(t) = g(X(t), \x_i(t)),
\label{eq:general_ode} 
\end{equation} 
where $g(X)$ is defined as
\begin{equation} 
    g(X) = 
    \begin{bmatrix} \sum_{i=1}^n \IL^{i}(\tau_m) \\ \x_c^{k+1}(\tau_m) - \Gamma(\x_i(t),\tau_m). \end{bmatrix} 
\end{equation} 
Here, $\tau_m$ represents the discretized time point indexed by $m$.
    Furthermore, we generalize the BE integration of the central agent ODE \eqref{eq:central_agent_be_step2} as:
\begin{equation}
    \rho(X^{k+1}(\tau_m)) = \Delta t \sigma( g(X^k(\tau_m), \Gamma(\x_1^{k+1}(T_1),\tau_m),\Gamma(\x_2^{k+1}(T_1),\tau_m),\cdots )),
\end{equation}
where the operator, $\rho(\cdot)$, is defined as
\begin{equation}
    \rho = \begin{bmatrix}
    1+\frac{\Delta t G_1^{th^{-1}}}{L} & 0 & \ldots & -\frac{\Delta t}{L} \\
         0 & 1+\frac{\Delta t G_2^{th^{-1}}}{L} & \ldots & -\frac{\Delta t}{L}\\
         0 & 0 & \ddots & \frac{-\Delta t}{L} \\
         -\Delta t  & -\Delta t  & \ldots & 1
    \end{bmatrix},
\end{equation}
and $\sigma(\cdot)$, is defined as
\begin{equation}
    \sigma = \begin{bmatrix}
       -\Gamma(\x_{1}^{k+1}(T_1),\tau_m) + \IL^{1^k}(\tau_m)G_1^{th^{-1}} \\
       -\Gamma(\x_{2}^{k+1}(T_2),\tau_m) + \IL^{2^k}(\tau_m)G_2^{th^{-1}} \\
        \vdots \\
        0
    \end{bmatrix}.
\end{equation}
Note, the operator, $\rho(\cdot)$, can be inverted to evaluate the central agent states:
\begin{equation}
    X^{k+1}(\tau) = \Delta t \rho^{-1} \sigma(g(X^k(\tau_m), \Gamma(\x_1^{k+1}(T_1),\tau_m),\Gamma(\x_2^{k+1}(T_1),\tau_m),\cdots )).
    \label{eq:multirate_proof1}
\end{equation}

The continuous-time ODE of \eqref{eq:multirate_central_ode_be},\eqref{eq:multirate_central_ode_be_step} converges to a stationary point characterized by $\x_c = \x_i\; \text{for all } i\in C$ and $\dot{I}_L^i=0$. The proof of convergence is provided in \cite{ecado}. In this analysis, we study the multi-rate discretization of the ODE to ensure the Gauss-Seidel process of solving the coupled system converges toward the steady state at each iteration. This proof uses the analysis for a multirate waveform relaxation for circuit simulation from \cite{white2012relaxation}. This analysis is based on proving that the central agent step is a contraction mapping towards the steady state (i.e., the stationary point of the global objective function).

The multi-rate integration uses the linear operator, $\Gamma(\cdot)$, to interpolate and extrapolate state variables at intermediate time points. Two important properties of the linear operator are the following:
\begin{enumerate}
    \item Given two signals, $\y(t)$ and $\z(t)$: 
    \begin{equation*}
    \Gamma(\y(t)+\z(t),\tau) = \Gamma(\y(t),\tau) + \Gamma(\z(t),\tau) 
    \end{equation*}
    
    \item Given a signal $\y(t)$, and a scalar, $\alpha$: 
    \begin{equation*}
        \Gamma(\alpha \y(t),\tau) = \alpha \Gamma(\y(t),\tau)
    \end{equation*}
    
\end{enumerate}

To prove convergence of the multi-rate integration step, we employ a continuous-time $\beta>0$ norm defined as:
\begin{equation}
    \|\y\|_{\beta} = \text{max}_{[0,T]} e^{-\beta t} [\text{max}_i \Gamma(\y_i(t),\tau) \;\; \forall i \in C]
\end{equation}

Under certain conditions, \eqref{eq:multirate_proof1} is a contraction mapping on the $\beta$ norm. To prove this relation, we evaluate the difference between two series, $\{X^k(\tau_m)\}$ and $\{Y^k(\tau_m)\}$, as follows:
\begin{multline}
    \{X^{k+1}(\tau_m)\} - \{Y^{k+1}(\tau_m)\} = \Delta t \rho^{-1}  \sigma(g(X^k(\tau_m), \Gamma(\x_1^{k+1}(T_1),\tau_m),\Gamma(\x_2^{k+1}(T_1),\tau_m),\cdots )) \\
    - \Delta t \rho^{-1} \sigma(g(Y^k(\tau_m), \Gamma(\y_1^{k+1}(T_1),\tau_m),\Gamma(\y_2^{k+1}(T_1),\tau_m),\cdots )).
\end{multline}
Exploiting the linearity of the operators, $\Gamma(\cdot)$ and $\rho$, leads to the following:
\begin{multline}
      \{X^{k+1}(\tau_m)\} - \{Y^{k+1}(\tau_m)\} = \Delta t \rho^{-1} \sigma[g(X^k(\tau_m), \Gamma(\x_1^{k+1}(T_1),\tau_m),\Gamma(\x_2^{k+1}(T_2),\tau_m),\cdots)  \\
      - g(Y^k(\tau_m), \Gamma(\y_1^{k+1}(T_1),\tau_m),\Gamma(\y_2^{k+1}(T_2),\tau_m),\cdots )].
\end{multline}

The BE operator $\rho^{-1}\sigma(\cdot)$ can be expanded into a series of summations, as shown in Appendix \ref{sec:expanding_be}, using the following equation: 
\begin{multline}
\{X^{k+1}(\tau_m)\} - \{Y^{k+1}(\tau_m)\} = \Delta t \sum_{l=0}^{m} \gamma_l[g(X^k(\tau), \Gamma(\x_1^{k+1}(T_1),\tau_{m-l}),\Gamma(\x_2^{k+1}(T_1),\tau),\cdots) 
\\ -g(Y^k(\tau), \Gamma(\y_1^{k+1}(T_1),\tau),\Gamma(\y_2^{k+1}(T_1),\tau),\cdots) ]
\label{eq:multirate_proof_be_operator}
\end{multline} 
where $\gamma_l$ is a scalar that determines the weight of the past state values in the numerical integration method.

To prove that the difference between the series is a contraction mapping, the following two lemmas are useful.

\begin{lemma}
    Given two sequences, \{$X(\tau)$\} and\{$Y(\tau)$\}, if $X(T_i)>Y(T_i)$ $\forall i$, then $\Gamma(X(T_i),\tau) > \Gamma(Y(T_i),\tau)$. Furthermore, if $X(T_i)=K \;\;\forall i$ where $K$ is a constant value, then $\Gamma(X(T_i),\tau)=K$. \label{lm:multirate_lemma1}
\end{lemma}

\begin{lemma}
\label{lm:multirate_lemma2}
    The $\beta$ norm on the following series of $X^K$ is bounded according to the following:
    \begin{equation}
    \label{eq:beta_norm}
        \max_{[0,T]}e^{-\beta \tau} \| \sum_{l=0}^{m}\|\gamma_l \Gamma(X,\tau_{m-l})\|\| \leq \frac{M}{1-e^{-\beta \Delta t}}e^{-\beta \tau} \|\Gamma (X,\tau_m) \|,
    \end{equation}
    where $M$ is equal to $\max_l \| \gamma_l\|$.
\end{lemma}

Proof of the two lemmas is provided in Appendixes \ref{sec:proof_multirate_lemma1} and \ref{sec:proof_multirate_lemma2}. Using Lemma \ref{lm:multirate_lemma1} and the Lipschitz constant of $\nabla f$, we can bound \eqref{eq:multirate_proof_be_operator} as follows:
\begin{equation}
     \{X^{k+1}(\tau)\} - \{Y^{k+1}(\tau)\} \leq  | \sum_{l=0}^{m} |\gamma_l| \left[ \sum_{j=1}^{i}\Delta t_i L_{ij} |\Gamma(X_j^{k+1}-Y_j^{k+1},\tau_{m-l})| + \sum_{j=i+1}^{n}\Delta t_i L_{ij}|\Gamma(X_j^{k}-Y_j^{k},\tau_{m-l})| \right],
\end{equation}
where $L_{ij}$ is the Lipschitz constant of the $i$th row of $g$ with respect to $X_j$. Using the triangle inequality, we formulate this as follows:
\begin{equation}
     \{X^{k+1}(\tau)\} - \{Y^{k+1}(\tau)\} \leq  |  \sum_{j=1}^{i}\Delta t_i L_{ij} \sum_{l=0}^{m} |\gamma_l| |\Gamma(X_j^{k+1}-Y_j^{k+1},\tau_{m-l})| + \sum_{j=i+1}^{n}\Delta t_i L_{ij}\sum_{l=0}^{m} |\gamma_l||\Gamma(X_j^{k}-Y_j^{k},\tau_{m-l})|.
\end{equation}
Multiplying both sides by $e^{-\beta t}$ and taking the maximum over the time window, $[0,T]$, results in the following:
\begin{multline}
     \max_{[0,T]} e^{-\beta t} \{X^{k+1}(\tau)\} - \{Y^{k+1}(\tau)\} \leq  |  \sum_{j=1}^{i}\Delta t_i L_{ij} \max_{[0,T]} e^{-\beta t} \sum_{l=0}^{m} |\gamma_l| |\Gamma(X_j^{k+1}-Y_j^{k+1},\tau_{m-l})| \\ 
     +  \sum_{j=i+1}^{n}\Delta t_i L_{ij} \max_{[0,T]} e^{-\beta t} \sum_{l=0}^{m} |\gamma_l||\Gamma(X_j^{k}-Y_j^{k},\tau_{m-l})|,
\end{multline}
where $T= \text{max}(T_i)$.
Using Lemma \ref{lm:multirate_lemma2}, we conclude the following:
\begin{multline}
     \max_{[0,T]} e^{-\beta t} \{X^{k+1}(\tau)\} - \{Y^{k+1}(\tau)\} \leq \left[\frac{M\Delta t_i}{1-e^{-\beta \Delta t_i}}\sum_{j=1}^{i}L_{ij} \right] \| X^{k+1} - Y^{k+1}\|_\beta \\
     + \left[\frac{M\Delta t_i}{1-e^{-\beta \Delta t_i}}\sum_{j=i+1}^{n}L_{ij} \right] \| X^{k} - Y^{k}\|_\beta ,
\end{multline}
where $\| \cdot\|_\beta$ is the $\beta$ norm defined in \eqref{eq:beta_norm}.

Assuming all time steps, $\Delta t_i>0$ are positive and a $\beta>0$, then there exists a scalar, $\delta>0$ such that:
\begin{equation}
    \delta > \frac{M\Delta t_i}{1-e^{-\beta \Delta t_i}}\sum_{j=i+1}^{n}L_{ij}.
\end{equation}
Using the definition of $\delta$, we conclude the following:
\begin{multline}
     \max_{[0,T]} e^{-\beta t} \{X^{k+1}(\tau)\} - \{Y^{k+1}(\tau)\} \leq \delta \| X^{k+1} - Y^{k+1}\|_\beta      + \delta \| X^{k} - Y^{k}\|_\beta.
\end{multline}
Because the value of $\delta$ holds for all time-steps (indexed by $i$), then:
\begin{multline}
     |\{X^{k+1}(\tau)\} - \{Y^{k+1}(\tau)\}|_\beta \leq \delta \| X^{k+1} - Y^{k+1}\|_\beta      + \delta \| X^{k} - Y^{k}\|_\beta,
\end{multline}
which can be rewritten as:
\begin{equation}
    \| X^{k+1}-Y^{k+1}\|_{\beta} \leq \frac{\delta}{1-\delta} \| X^{k}-Y^{k}\|_{\beta}.
\end{equation}
This proves that for a value of $\delta$ such that $\frac{\delta}{1-\delta}<1$, the multirate integration scheme, $\rho^{-1}\sigma(\cdot)$, is a contraction mapping whereby the series $\{X^{k+1}(\tau)\}$ and $\{Y^{k+1}(\tau)\}$ converges to a stationary point.

Note, the rate of the contraction mapping, determined by $\frac{\delta}{1-\delta}$, is not affected by the ratio of client step-sizes, $\Delta t_i$. This enables clients to take vastly different step sizes, with accuracy considerations imposed by a Local Truncation Error (LTE). The LTE estimates the error in the approximation of the numerical integration and is a key measure of the accuracy at any iteration.

To prove convergence to a stationary point of the global objective function, we consider the contraction mapping between two series, $\{X^k\}$ and $\{X^*\}$, where $X^*$ is the state at the stationary point (i.e., $g(X^*)=0$). The difference between the series is diminished according to the contraction mapping:
\begin{equation}
    \|X^{k+1} - X^*\|_{\beta} \leq \frac{\delta}{1-\delta}\|X^{k} - X^*\|_{\beta}.
\end{equation}

This implies that for a $\Delta t$ ensuring $\frac{\delta}{1-\delta}<1$, the FedECADO update asymptotically converges to a stationary point, with a convergence rate determined by $\frac{\delta}{1-\delta}$.

\end{proof}

\subsection{Expanding Backward Euler Operator}
\label{sec:expanding_be}

Given a general ODE
\begin{equation}
    \dot{z}(t) = s(z,t)
\end{equation}
with an initial known state of $z(0)$, the state of $z$ at a time point, $\tau_m$ is determined by solving the following:
\begin{equation}
    z(\tau_m) = z(\tau_{m-1}) + \int_{\tau_{m-1}}^{\tau_m} s(z(\tau),\tau) d \tau.
\end{equation}
The integral on the right-hand side generally does not have a closed form solution and is approximated using a generalized numerical integration method:
\begin{equation}
    z(\tau_m) = z(\tau_{m-1}) + \sum_{l=0}^{n} k_l s(\tau_l, \tau_l),
\end{equation}
where $k_l$ is a scalar used to weight the contribution of the state at time $\tau_l$. This expression can be further generalized:
\begin{equation}
    z(\tau_m) = \sum_{l=0}^{m} \gamma_l s(\tau_l, \tau_l) + z(0),
\end{equation}
where $\gamma_l$ weights the contribution of past states.

We can apply this form for the Backward-Euler integration step of the ODE \eqref{eq:general_ode}, which is generally written as follows:
\begin{equation} X(\tau_m) = X(\tau_{m-1}) + \Delta t_m g(X(\tau_{m},x_i(\tau_m)), \end{equation} 
where the index $m$ represents the iteration of Backward Euler steps taken. An equivalent representation is the following: \begin{equation} X(\tau_m) = \sum_{j=0}^{m} \Delta t_{m-j} g(X(\tau_{m-j},x_i(\tau_{m-j})) + X(0). 
\end{equation}
This expresses the latest state $X(\tau_m)$ as a summation of previous values of $g(\cdot)$.

\subsection{Proof of Lemma \ref{lm:multirate_lemma1}}
\label{sec:proof_multirate_lemma1}
The proof of Lemma \ref{lm:multirate_lemma1} is a direct consequence of the linearity of the operator, $\Gamma$. Consider two sequences, ${X(\tau)}$ and ${Y(\tau)}$, which are evaluated at time points $t_1$ and $t_2$, where by:
\begin{align}
    X(t_1) > Y(t_1) \\
    X(t_2) > Y(t_2).
\end{align}
Applying the linear operator, $\Gamma(\cdot, \tau)$ for a time point $\tau \in [t_1,t_2]$, is defined as follows:
\begin{align}
    \Gamma(X,\tau) = \frac{X(t_2)-X(t_1)}{t_2 - t_1}(\tau-t_1) + X(t_1) \\ 
    \Gamma(Y,\tau) = \frac{Y(t_2)-Y(t_1)}{t_2 - t_1}(\tau-t_1) + Y(t_1)
\end{align}
Because $X > Y$ at time points $t_1$ and $t_2$, we observe the following:
\begin{equation}
    \frac{X(t_2)-X(t_1)}{t_2 - t_1}(\tau-t_1) + X(t_1) > \frac{Y(t_2)-Y(t_1)}{t_2 - t_1}(\tau-t_1) + Y(t_1)
\end{equation}
thereby proving that $\Gamma(X,\tau) > \Gamma(Y,\tau)$.
Expanding this proof to multiple evaluated time points $T_i$, we note that if $X(T_i)>Y(T_i)\forall i$, then $\Gamma(X,\tau) > \Gamma(Y,\tau)$.

\subsection{Proof for Lemma \ref{lm:multirate_lemma2}}
\label{sec:proof_multirate_lemma2}

The proof of Lemma \ref{lm:multirate_lemma2} is reconstructed in the following from [\cite{white2012relaxation}].

From the definition of the $\beta$-norm, we see that:
\begin{equation}
    \| \sum_{l=0}^{m} \gamma_l X(\tau_{m-l}) \|_{\beta} = \max_m e^{-\beta \Delta t m} \| \sum_{l=0}^{m} \gamma_l X(\tau_{m-l}) \|,
\end{equation}
which can be upper-bounded using the triangle inequality as:
\begin{equation}
    \| \sum_{l=0}^{m} \gamma_l X(\tau_{m-l}) \|_{\beta} \leq \max_m e^{-\beta \Delta t m} \sum_{l=0}^{m} |\gamma_l| \| X(\tau_{m-l}) \|.
\end{equation}

Multiplying $e^{\beta (m-l)\Delta t}e^{-\beta (m-l)\Delta t}$ (equal to 1) into the right-hand side of the equation above leads to
\begin{equation}
        \| \sum_{l=0}^{m} \gamma_l X(\tau_{m-l}) \|_{\beta} \leq \max_m e^{-\beta \Delta t m} \sum_{l=0}^{m} |\gamma_l| e^{\beta (m-l)\Delta t}e^{-\beta (m-l)\Delta t}\| X(\tau_{m-l}) \|.
\end{equation}
Because $e^{-\beta (m-l)\Delta t}\| X(\tau_{m-l}) \| \leq \| X(\tau_{m-l}\|_{\beta}$, then 
\begin{equation}
      \| \sum_{l=0}^{m} \gamma_l X(\tau_{m-l}) \|_{\beta} \leq \max_m e^{-\beta \Delta t m} \sum_{l=0}^{m} |\gamma_l| \| X(\tau_{m-l}) \|_{\beta}.
\end{equation}

Suppose $|\gamma_l|$ is upper-bounded by $M$, then the inequality becomes
\begin{equation}
      \| \sum_{l=0}^{m} \gamma_l X(\tau_{m-l}) \|_{\beta} \leq M \sum_{l=0}^{m} [e^{-\beta \Delta t m}]  \| X(\tau_{m-l}) \|_{\beta}
\end{equation}
Because $e^{-\beta \Delta m}>0$ and $\sum_{l=0}^{m}e^{-\beta \Delta m} \leq\sum_{l=0}^{\infty}e^{-\beta \Delta m} $, then the inequality is as follows:
\begin{equation}
      \| \sum_{l=0}^{m} \gamma_l X(\tau_{m-l}) \|_{\beta} \leq M \sum_{l=0}^{\infty} [e^{-\beta \Delta t m}]  \| X(\tau_{m-l}) \|_{\beta}.
\end{equation}
The infinite series can be directly computed as:
\begin{equation}
    \sum_{l=0}^{\infty}e^{-\beta \Delta m} = \frac{e^{-\beta \Delta t}}{1-e^{-\beta \Delta t}},
\end{equation}
where the upper bound is as follows:
\begin{equation}
      \| \sum_{l=0}^{m} \gamma_l X(\tau_{m-l}) \|_{\beta} \leq M \frac{e^{-\beta \Delta t}}{1-e^{-\beta \Delta t}}e^{-\beta \Delta t m}] \| X(\tau_{m-l}) \|_{\beta},
\end{equation}
which proves Lemma \ref{lm:multirate_lemma2}.

\section{FedECADO Algorithm}
\label{sec:fedecado_alg}
The full workflow for the FedECADO algorithm is shown in Algorithm \ref{fedecado_alg}. The algorithm begins by initializing the state values in Steps 1–4 and precomputing the constant sensitivity model, $\bar{G}_{th}$, for all clients in Step 5. The hyperparameters for the algorithm are $L>0$ and the local truncation error tolerance, $\eta>0$.

For each epoch, FedECADO begins by simulating the set of active clients, $C_a \in C$, in step 10, for a number of epochs, $e_i$. The client ODE is solved by using numerical integration selected by the user (further details are provided in \cite{agarwal2023equivalent}). In step 11 of Algorithm \ref{fedecado_alg}, we use a Forward-Euler integration to simulate the local ODE. 

The active clients then communicate their final states, $\x_i^{k+1}(t+T_i)$, and simulation time, $T_i$, to the central agent server, which evaluates its own states at intermediate time points (steps 12-16). First the central agent estimates the active client state values at a time point, $\tau$, using the operator, $\Gamma(\cdot, \tau)$. Then after selecting a time step, $\Delta t$, that satisfies the accuracy conditions in \eqref{eq:be_lte_cap},\eqref{eq:be_lte_ind}, the central agent solves for its states at the proceeding time point, $\tau+\Delta t$, in step 15. The central agent server progresses through time (performing steps 13-14), until it has simulated the maximum client simulation time window determined by $\max(T_i)$. 

Note, for a given time-step, $\Delta t$, the central agent LU-factorizes the left hand matrix in Step 16 of Algorithm \ref{fedecado_alg} \eqref{eq:central_agent_be_step2}, so that any subsequent central agent steps with the same step size only requires a forward-backward substitution to solve for the central agent states. In practice, once an appropriate step size, $\Delta t$, is selected that satisfies the accuracy conditions in \eqref{eq:be_lte_cap}-\eqref{eq:be_lte_ind}, it is infrequently updated. As a result, re-using the same LU-factor provides a computational advantage across multiple central agent evaluations and improves the overall runtime performance of FedECADO.

\begin{algorithm}
    \caption{FedECADO Central Update}
    \label{fedecado_alg}
    \textbf{Input: } $\nabla f(\cdot)$,$\x(0)$, $\eta>0, L>0$
    
    \begin{algorithmic}[1]
    \STATE{$\x_c \gets \x(0)$}
    \STATE{$\x_i \gets \x(0)$}
    \STATE{$I_i^L \gets 0$}
    \STATE{$t \gets 0$}

    \STATE{Precompute $\bar{G}_i^{th} \; \forall i\in C$} 
    
    \STATE{\textbf{do while} $\|\dot{\x}_c\|^2 > 0$}
    \STATE{\hspace*{\algorithmicindent}$\x_c^k \gets \x_c^{k+1}$}
    \STATE{\hspace*{\algorithmicindent}$\x_i^{k} \gets \x^{i^{k+1}}$}
    
    \STATE{\hspace*{\algorithmicindent}\textit{Parallel Solve for active client states, $\x_i^{k+1}(t+T_i) \forall i\in C_a$, by simulating:} }

   \STATE{\hspace*{\algorithmicindent} \hspace*{\algorithmicindent}\textbf{for $e_i$ epochs:}}
    
    \STATE{\hspace*{\algorithmicindent}\hspace*{\algorithmicindent}\hspace*{\algorithmicindent}$\x_i^{k+1}(t+\Delta t_i) = \x_i^{k+1}(t) - \Delta t_i\nabla f(\x_i^{k+1}(t)) - \Delta t_i I_i^{L^k}(t) $}


    \STATE{\hspace*{\algorithmicindent} \textbf{for $\tau \in[t,t+\max(T_i)]$}}
    
    \STATE{\hspace*{\algorithmicindent} \hspace*{\algorithmicindent} Select $\Delta t$ according to Algorithm \ref{adaptive-time-step}}
     \STATE{\hspace*{\algorithmicindent}\hspace*{\algorithmicindent}  Evaluate active client states at timepoint $\tau$: $\Gamma(x_i^{k+1},\tau) \;\;\forall i \in C_a$}
         \STATE{\hspace*{\algorithmicindent}\hspace*{\algorithmicindent}\textit{Solve for $\x_c^{k+1}(\tau+\Delta t), I_i^{L^{k+1}}(\tau+\Delta t)$} according to (\ref{eq:central_agent_be_step2})}

    \STATE{\hspace*{\algorithmicindent}\hspace*{\algorithmicindent} $\tau = \tau+\Delta t$}
    \RETURN $\x_c$
    \end{algorithmic}
    \end{algorithm}

\section{Scaling the FedECADO Algorithm}
\label{sec:scaling_fedecado}
The previous experiments highlight the individual contribution of the proposed aggregate sensitivity model and multi-rate integration on distributed training. In this experiment, we study the impact of both methods to address the challenges of federated learning. We train a larger ResNet-34 model on a CIFAR-100 dataset distributed across 100 clients with both non-IID data distribution as well as heterogeneous learning rates and numbers of epochs, determined by \eqref{eq:random_lr},\eqref{eq:random_epochs}.

\begin{figure}
    \centering
    \begin{subfigure}{0.45\columnwidth}
        \centering
        \includegraphics[width=0.9\columnwidth,]{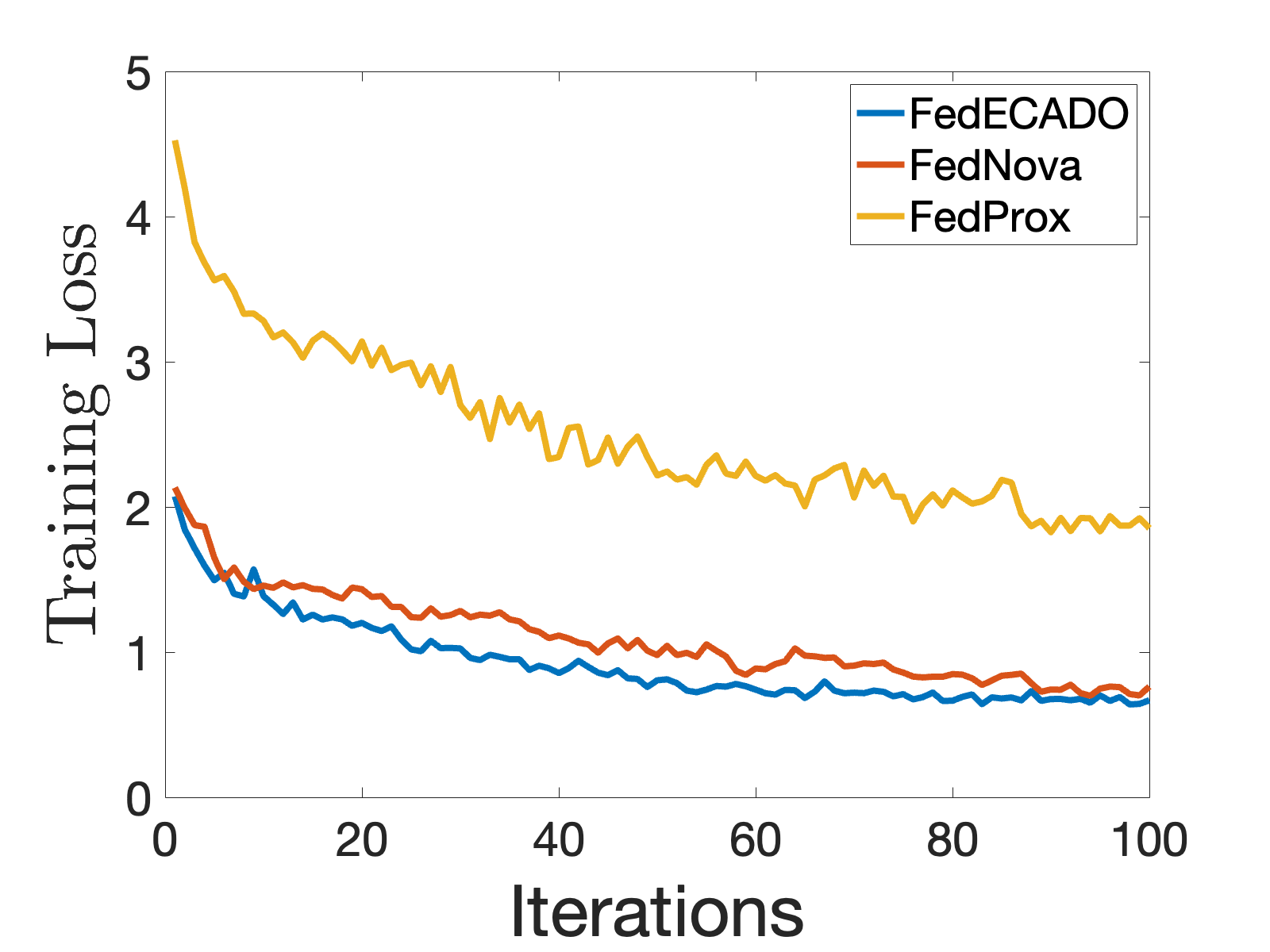}
    \end{subfigure}%
    \hfill
    \begin{subfigure}{0.45\columnwidth}
        \centering
        \includegraphics[width=0.9\columnwidth]{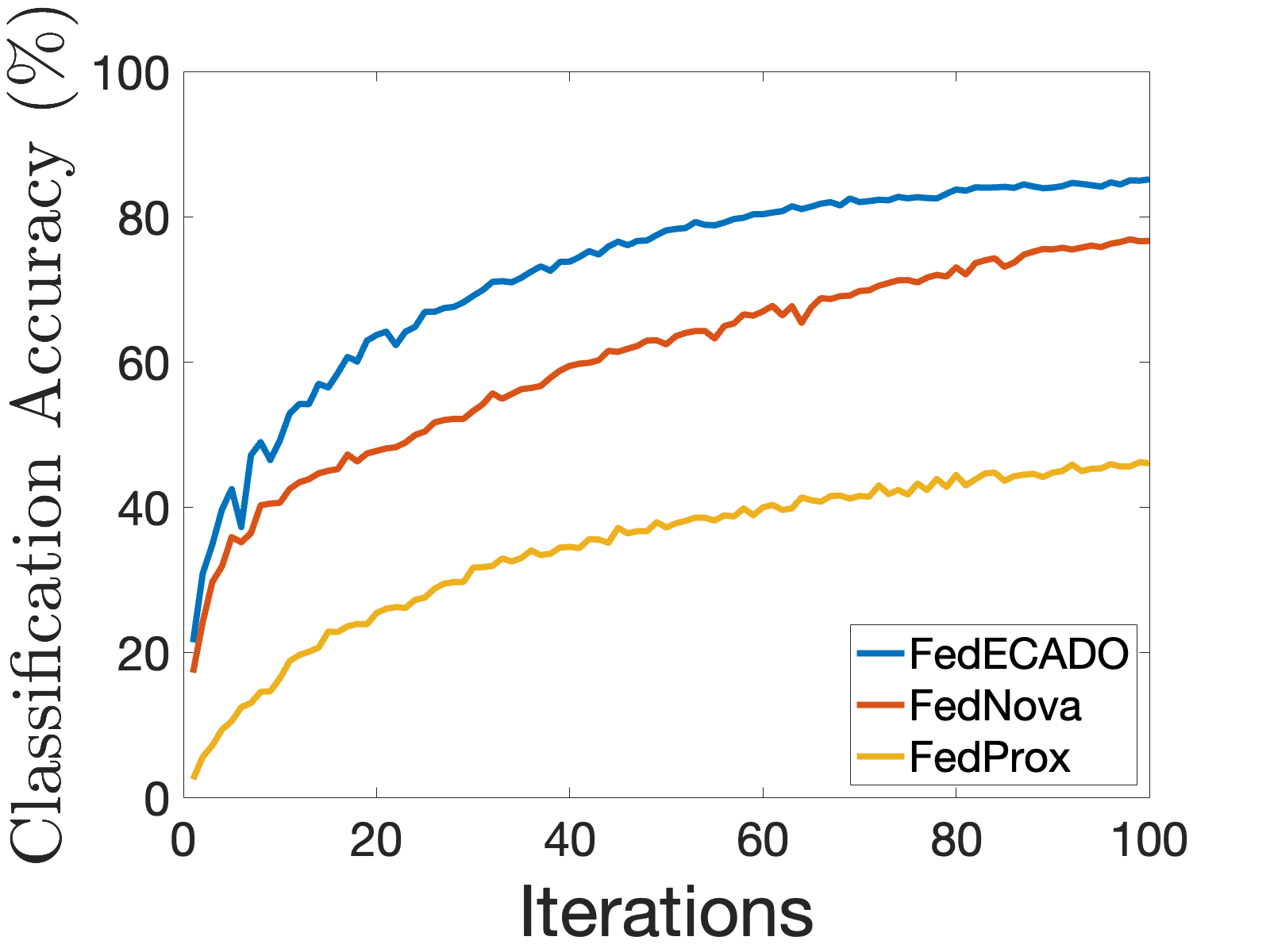}
    \end{subfigure}
    \caption{Scaling FedECADO to train ResNet34 model on CIFAR-100 dataset distributed on 100 clients with heterogeneous computation (non-IID data distribution and asynchronous updates.}
    \label{fig:resnet_experiment}
    \end{figure}

Figure \ref{fig:resnet_experiment} showcases FedECADO's advantage over FedNova and FedProx in training the ResNet-34 model. Our approach achieves a lower training loss, indicating more efficient convergence, as well as higher classification accuracy after 100 epochs (4.6\% higher than FedNova and 8.6\% higher than FedProx).

Despite evaluating the central agent multiple times for multi-rate integration, FedECADO’s clock time for centralized updates is comparable to those of FedNova and FedProx (only 1\% slower than FedNova and 2.4\% slower than FedProx). 
The main computational cost of Algorithm \ref{fedecado_alg} occurs on the subset of active clients during each communication round, which is significantly smaller than the total client base. The resulting BE matrix \eqref{eq:central_agent_be_step2} is relatively small, with dimensions of $\mathcal{R}^{a\times a}$, where $a$ is the size of the active client list. Furthermore, pre-computing the LU matrix of in \eqref{eq:central_agent_be_step2} minimizes the total computation time for central agent evaluations.

\end{document}